\documentclass[final]{cvpr}

\usepackage{times}
\usepackage{epsfig}
\usepackage{graphicx}
\usepackage{amsmath}
\usepackage{amssymb}

\usepackage[pagebackref=true,breaklinks=true,colorlinks,bookmarks=false]{hyperref}

\usepackage[utf8]{inputenc} 
\usepackage[T1]{fontenc}    
\usepackage{url}            
\usepackage{booktabs}       
\usepackage{amsfonts}       
\usepackage{nicefrac}       
\usepackage{microtype}      

\usepackage{xcolor}

\usepackage{caption}
\usepackage{subcaption}
\usepackage{floatrow}

\usepackage[flushleft]{threeparttable} 
\usepackage{cite}
\usepackage{multirow}
\usepackage{tabularx}
\usepackage{booktabs}
\usepackage[export]{adjustbox}

\newsavebox{\measurebox}

\def\figvspace{{\vspace{-3mm}}}
\def\tablespace{{\vspace{-3mm}}}

\newcommand{\Section}[1]{\vspace{-0mm} \section{#1} \vspace{0mm}}
\newcommand{\SubSection}[1]{\vspace{-1mm} \subsection{#1} \vspace{-1mm}}

\def\cvprPaperID{9944} 
\def\confYear{CVPR 2021}

\begin{document}

\title{Mitigating Face Recognition Bias via Group Adaptive Classifier}

\author{Sixue Gong\quad\quad Xiaoming Liu\quad\quad Anil K. Jain\\
Michigan State University, East Lansing MI 48824\\
{\tt\small \{gongsixu, liuxm, jain\}@msu.edu}
}

\maketitle

\begin{abstract}
  Face recognition is known to exhibit bias - subjects in a certain demographic group can be better recognized than other groups. 
  This work aims to learn a fair face representation, where faces of every group could be more equally represented. 
  Our proposed group adaptive classifier mitigates bias by using adaptive convolution kernels and attention mechanisms on faces based on their demographic attributes. 
  The adaptive module comprises kernel masks and channel-wise attention maps for each demographic group so as to activate different facial regions for identification, leading to more discriminative features pertinent to their demographics. 
Our introduced automated adaptation strategy  determines whether to apply adaptation to a certain layer by iteratively computing the dissimilarity among demographic-adaptive parameters. 
  A new de-biasing loss function is proposed to mitigate the gap of average intra-class distance between demographic groups.
  Experiments on face benchmarks (RFW, LFW, IJB-A, and IJB-C) show that our work is able to mitigate  face recognition bias across demographic groups while maintaining the competitive accuracy.
\end{abstract}

\section{Introduction}
Face recognition (FR) systems are known to exhibit discriminatory behaviors against certain demographic groups~\cite{howard2019effect, klare2012face, grother2019frvt}. The $2019$ NIST Face Recognition Vendor Test~\cite{grother2019frvt} shows that all $106$ tested FR algorithms exhibit varying biased performances on gender, race, and age groups of a mugshot dataset. 
Deploying biased FR systems to law enforcement is potentially unethical~\cite{creager2019flexibly}. 
Given the implication of automated FR-driven decisions, it is crucial to develop fair and unbiased FR systems to avoid the negative societal impact.
Note that, differ from the inductive bias in machine learning~\cite{dietterich1995machine}, we define FR bias as the {\it uneven} recognition performance w.r.t.~demographic groups. 

\begin{figure}[t]
    \captionsetup{font=footnotesize}
    \centering
    \begin{subfigure}[b]{0.68\linewidth}
    \includegraphics[width=\linewidth]{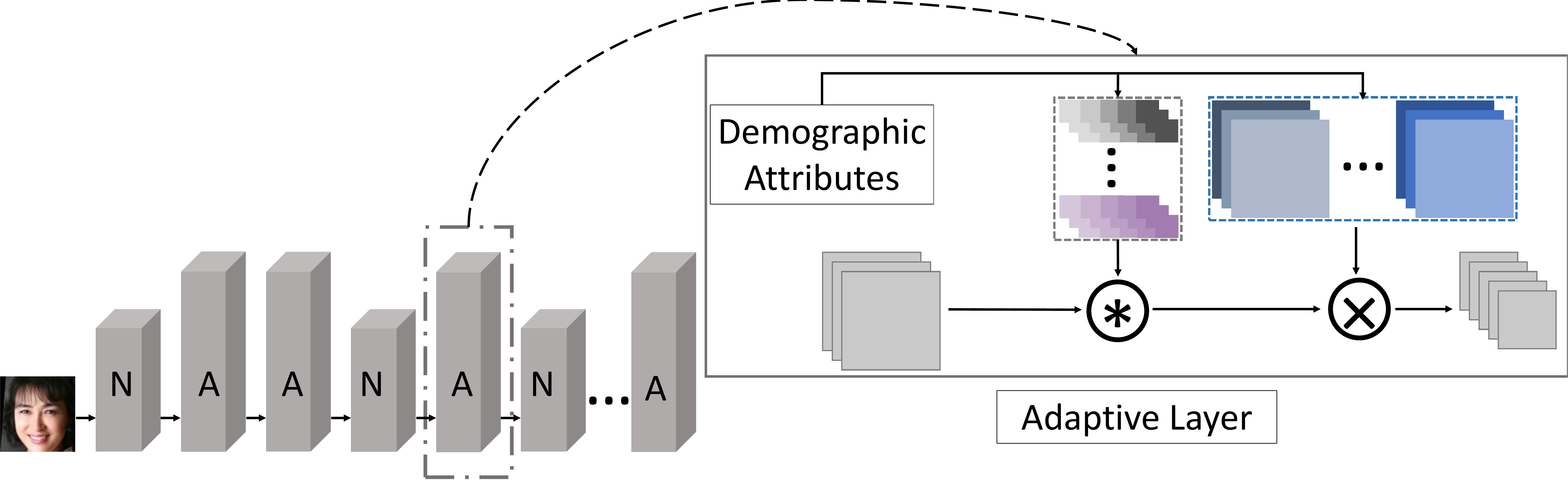}
    \vspace{-6mm}
    \caption{{\footnotesize}}
    \label{fig:fig1a}
    \end{subfigure}\hfill
    \begin{subfigure}[b]{0.31\linewidth}
    \includegraphics[width=0.9\linewidth]{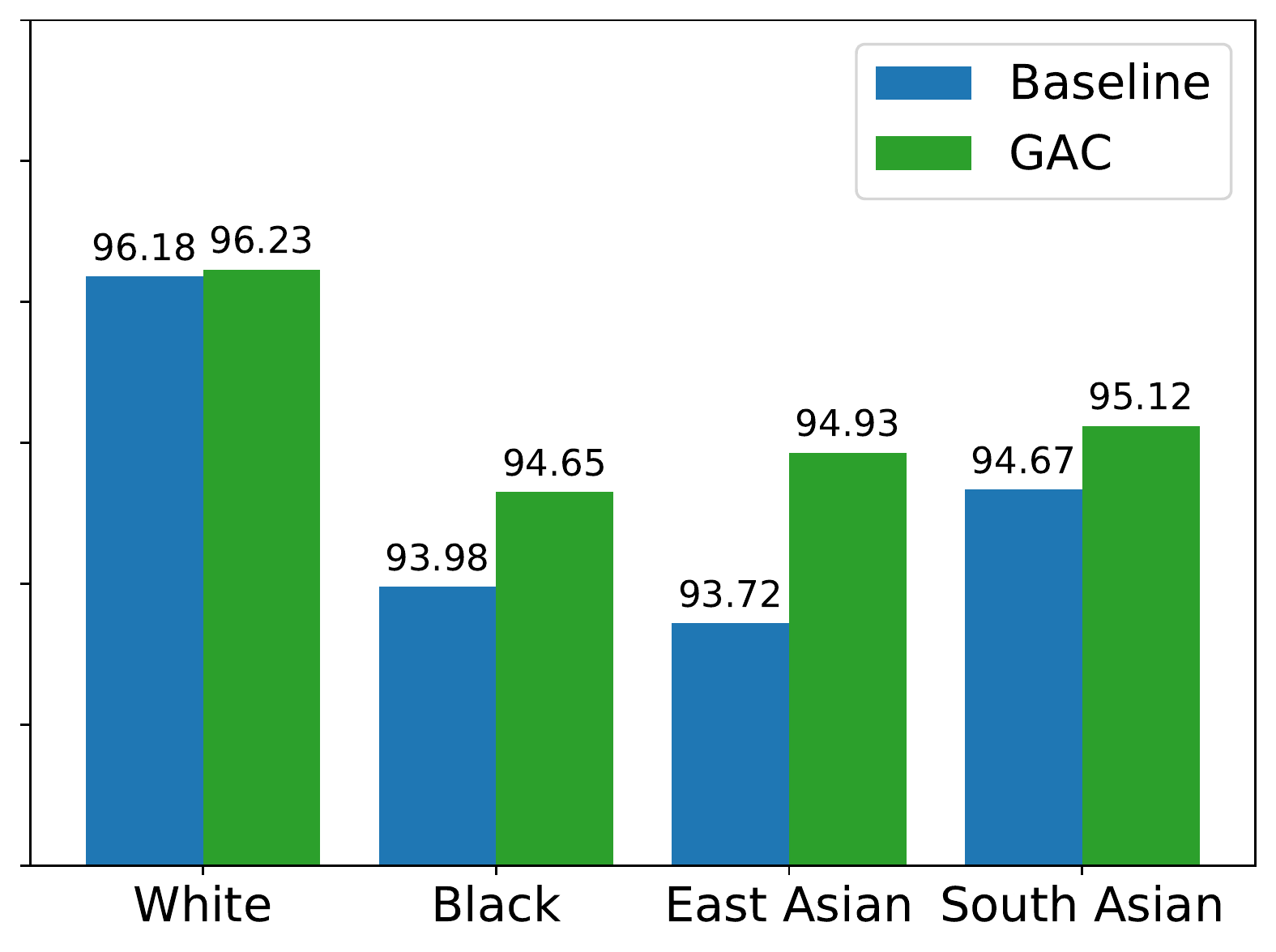}
     \vspace{-1mm}
    \caption{{\footnotesize}} 
    \label{fig:fig1b}
    \end{subfigure}\\
    \vspace{-1mm}
    \caption{\footnotesize{(a) Our proposed group adaptive classifier (GAC) automatically chooses between non-adaptive (``N'') and adaptive (``A'') layer in a multi-layer network, where the latter uses demographic-group-specific kernel and attention. (b) Compared to the baseline with the $50$-layer ArcFace backbone, GAC improves face verification accuracy in most groups of RFW dataset~\cite{wang2019racial}, especially under-represented groups, leading to mitigated FR bias, {\it e.g.}, GAC reduces biasness from $1.11$ to $0.60$.} \vspace{-4mm}}
    \label{fig:teaser}
\end{figure}

State-of-the-art (SOTA) FR algorithms~\cite{liu2017sphereface, wang2018cosface, Deng_2019_CVPR} rely on convolutional neural networks (CNNs) trained on large-scale face datasets. 
The public FR training datasets, {\it e.g.}, CASIA-WebFace~\cite{yi2014learning}, VGGFace2~\cite{cao2018vggface2}, and MS-Celeb-1M~\cite{guo2016ms}, are collected by scraping images off the web, with inevitable demographic bias~\cite{wang2020mitigating}. 
Biases in data  are transmitted to the FR models through network learning. For example, to minimize the overall loss, a network tends to learn a better representation for faces in the majority group whose number of faces dominate the training set, resulting in unequal discriminabilities. 
The imbalanced demographic distribution of face data is, nevertheless, not the only trigger of FR bias. 
Prior works have shown that even using a demographic balanced dataset~\cite{wang2020mitigating} or training separate classifiers for each group~\cite{klare2012face}, the performance on some groups is still inferior to the others.
By studying non-trainable FR algorithms, \cite{klare2012face} introduced the notion of \textit{inherent bias}, {\it i.e.}, certain groups are inherently more susceptible to errors in the face matching.

To tackle the dataset-induced bias, traditional methods re-weight either the data proportions~\cite{chawla2002smote} or cost values~\cite{akbani2004applying}. 
Such methods are limited when applied to large-scale imbalanced datasets. 
Recent imbalance learning methods focus on novel objective functions for class-skewed datasets. 
For instance, Dong~\etal~\cite{dong2018imbalanced} propose a Class Rectification Loss to incrementally optimize on hard samples of the classes with under-represented attributes. 
Alternatively, researchers strengthen the decision boundary to impede perturbation from other classes by enforcing margins between hard clusters via adaptive clustering~\cite{huang2019deep}, or between rare classes via Bayesian uncertainty estimates~\cite{khan2019striking}. 
To adapt the aforementioned methods to racial bias mitigation, Wang~\etal~\cite{wang2020mitigating} modify the large margin based loss functions by  reinforcement learning.
However,~\cite{wang2020mitigating} requires two auxiliary networks, an offline sampling network and a deep Q-learning network, to generate adaptive margin policy for training the FR network, which hinders the learning efficiency.

To mitigate FR bias, our main idea is to optimize the face representation learning on every demographic group in a single network, despite demographically imbalanced training data.
Conceptually, we may categorize face features into two types of patterns: \textit{general pattern} is shared by all faces; \textit{differential pattern} is relevant to demographic attributes.
When the differential pattern of one specific demographic group dominates training data, the network learns to predict identities mainly based on that pattern as it is more convenient to minimize the loss than using other patterns, leading to bias towards that specific group. 
One mitigation is to give the network more capacity to broaden its scope for multiple face patterns from different  groups. 
An unbiased FR model shall rely on both unique patterns for recognition of different groups, and general patterns of all faces for improved generalizability. 
Accordingly, as in Fig.~\ref{fig:teaser}, we propose a \textit{group adaptive classifier} (GAC) to explicitly learn these different feature patterns.
GAC includes two modules: the adaptive layer and automation module.
The adaptive layer comprises adaptive convolution kernels and channel-wise attention maps where each kernel and map tackle faces in {\it one} demographic group.
We also introduce a new objective function to GAC, which diminishes the variation of average intra-class distance between demographic groups.

\begin{figure*}[t!]
	    \captionsetup{font=footnotesize}
	    \centering
	    \begin{subfigure}[b]{0.65\linewidth}
	    \includegraphics[width=\linewidth]{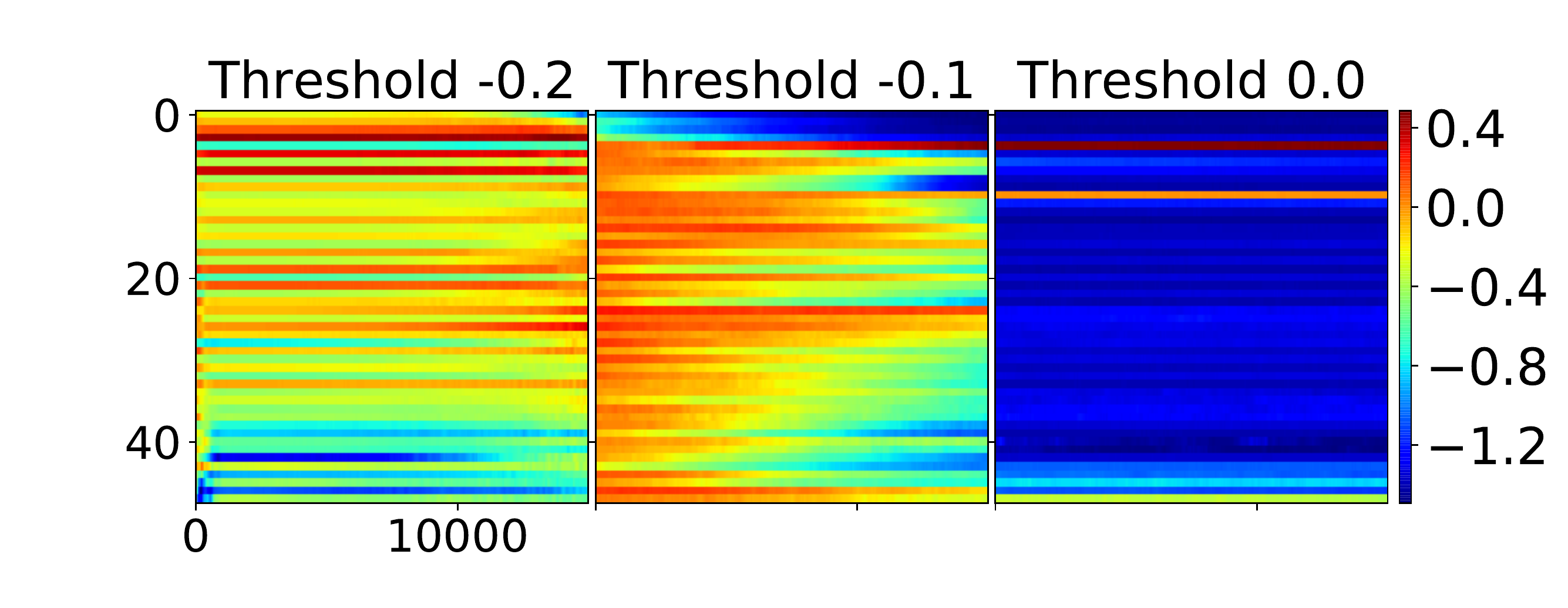}
	    \vspace{-5mm}
	    \caption{{\footnotesize}}
	    \label{fig:kernel_step}
	    \end{subfigure}\hfill
	    \begin{subfigure}[b]{0.32\linewidth}
	    \includegraphics[width=\linewidth]{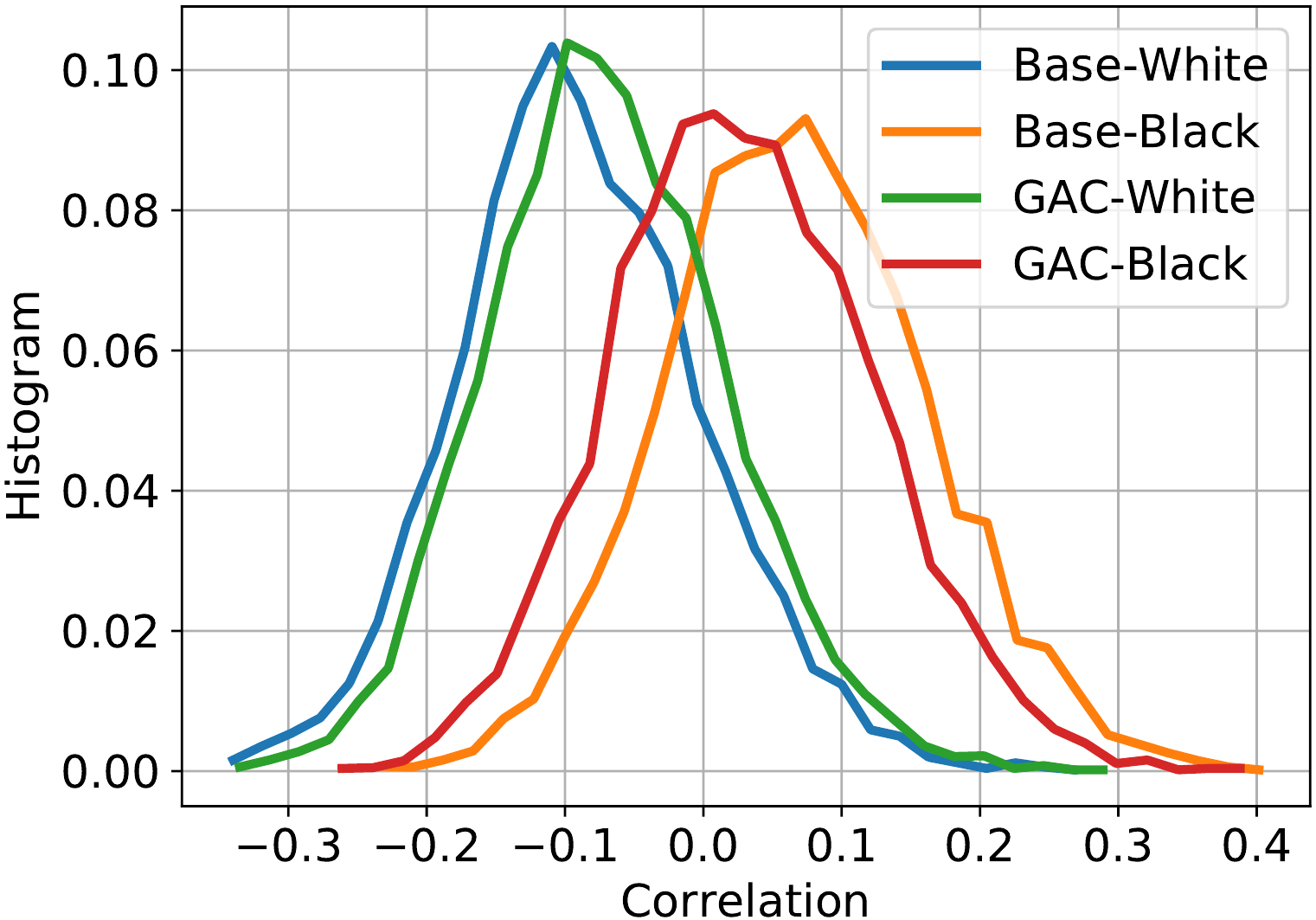}
	     \vspace{-5mm}
	    \caption{{\footnotesize}} 
	    \label{fig:pcso_histogram}
	    \end{subfigure}\\
	    \vspace{-3mm}
	    \caption{\footnotesize{ (a) For each of the three $\tau$ in automatic adaptation, we show the average similarities of pair-wise demographic kernel masks, {\it i.e.}, $\overline{\theta}$, at $1$-$48$ layers (y-axis), and $1$-$20K$ training steps (x-axis). The number of adaptive layers in three cases, {\it i.e.}, $\sum_{1}^{48}(\overline{\theta}>\tau)$ at $20K^{{th}}$ step, are $12$, $8$, and $6$, respectively.
	    (b) With two race groups (White, Black in PCSO~\cite{klare2012face}) and two models (baseline, GAC), for each of the four combinations, we compute pair-wise correlation of face representations using any two of $1$K subjects in the same race, and plot the histogram of correlations. GAC reduces the difference/bias of two distributions.    } \vspace{-2mm}}
	    \label{fig:heatmap}
	    
\end{figure*}

Prior work on dynamic CNNs introduce adaptive convolutions to either every layer~\cite{kang2017incorporating, yang2019cross, wang2019eca}, or manually specified layers~\cite{lu2019see, hou2019cross, su2019multi}. In contrast, we propose an automation module to choose which layers to apply adaptations. 
As we observed, not all convolutional layers require adaptive kernels for bias mitigation (see Fig.~\ref{fig:kernel_step}). 
At any layer of GAC, only kernels expressing high dissimilarity are considered as demographic-adaptive kernels. 
For those with low dissimilarity, their average kernel is shared by all inputs in that layer. 
Thus, the proposed network progressively learns to select the optimal structure for the demographic-adaptive learning. 
Both non-adaptive layers with shared kernels and adaptive layers are jointly learned in a unified network.

The contributions of this work include: 1) A new face recognition algorithm that reduces demographic bias and tailors representations for faces in every demographic group by adopting adaptive convolutions and attention techniques; 2) A new adaptation mechanism that automatically determines the layers to employ dynamic kernels and attention maps; 3) The proposed method achieves SOTA performance on a demographic-balanced dataset and three benchmarks.

\Section{Related Work}
\textbf{Fairness Learning and De-biasing Algorithms.} A variety of fairness techniques are proposed to prevent machine learning models from utilizing statistical bias in training data, including adversarial training~\cite{alvi2018turning, hendricks2018women, wang2019balanced, pmlr-v80-madras18a}, subgroup constraint optimization~\cite{kearns2019empirical, zhao2017men, wang2019towards}, data pre-processing ({\it e.g.}, weighted sampling~\cite{grover2019bias}, and data transformation~\cite{calmon2017optimized}), and algorithm post-processing~\cite{kim2019multiaccuracy, pleiss2017fairness}. 
Another promising approach learns a fair representation to preserve all discerning information about the data attributes or task-related attributes but eliminate the prejudicial effects from sensitive factors~\cite{moyer2018invariant, song2019learning, zemel2013learning, creager2019flexibly, hardt2016equality}. Locatello~\etal~\cite{locatello2019fairness} show the feature disentanglement is consistently correlated with increasing fairness of general purpose representations by analyzing $12,600$ SOTA models. 
Accordingly, a disentangled representation is learned to de-bias both FR and demographic attribute estimation~\cite{gong2020jointly}. Other studies address the bias issue in FR by leveraging unlabeled faces to improve the performance in minority groups~\cite{qin2020asymmetric, wang2019racial}. Wang~\etal~\cite{wang2020mitigating} propose skewness-aware reinforcement learning to mitigate racial bias.
Unlike prior work, our GAC is designed to customize the classifier for each  demographic group, which, if successful, would lead to mitigated bias.
\begin{figure*}[t!]
	    \captionsetup{font=footnotesize}
	    \centering
	    \begin{subfigure}[b]{0.27\linewidth}
	    \includegraphics[width=0.95\linewidth]{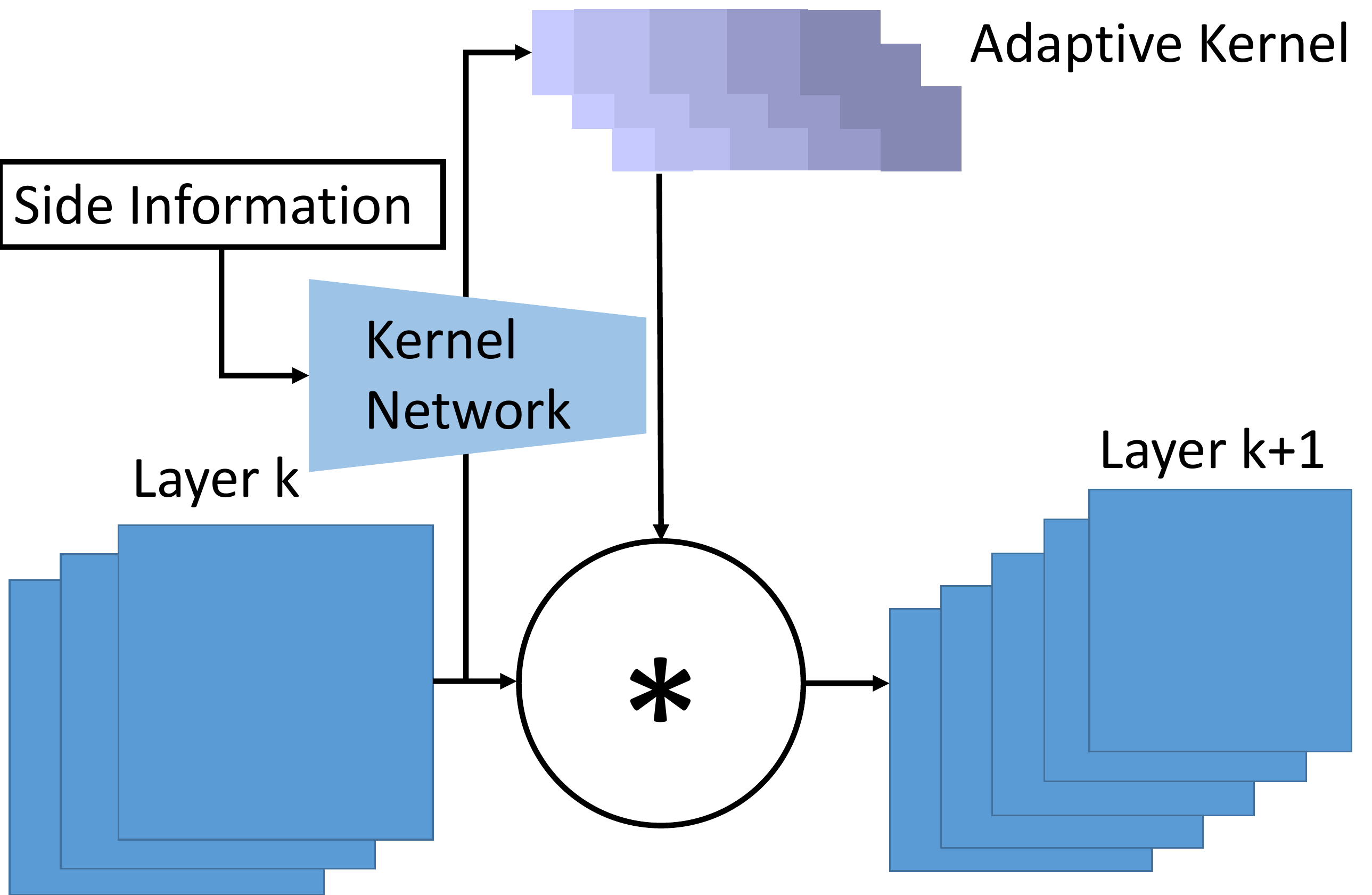}
	    \caption{{\footnotesize Adaptive Kernel}}
	    \label{fig:adap_kernel}
	    \end{subfigure}\hfill		
		\begin{subfigure}[b]{0.27\linewidth}
		\includegraphics[width=0.95\linewidth]{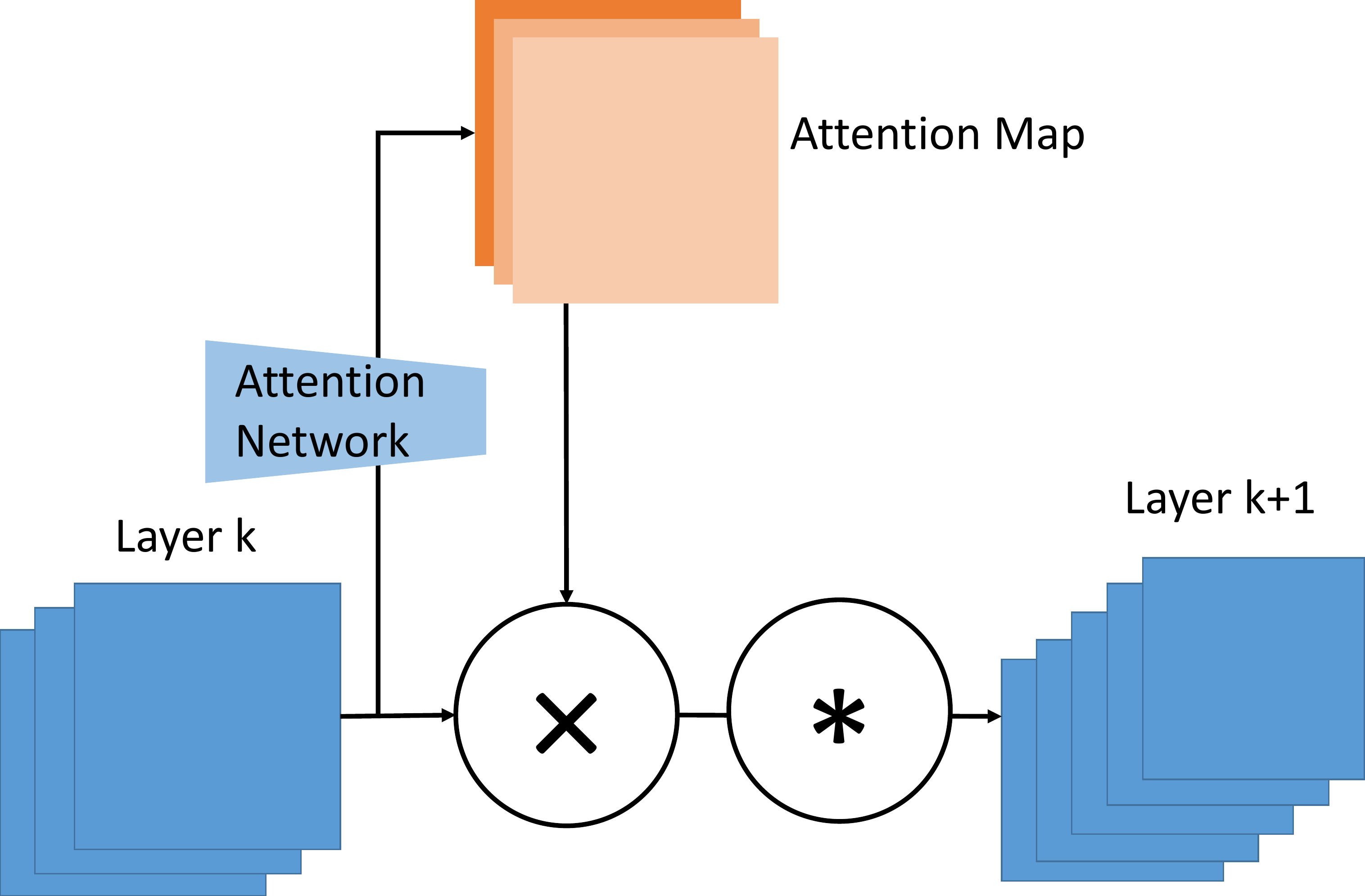}
	    \caption{{\footnotesize Attention Map}}
	    \label{fig:attention_map}
	    \end{subfigure}\hfill    
	    \begin{subfigure}[b]{0.45\linewidth}
	    \includegraphics[width=0.95\linewidth]{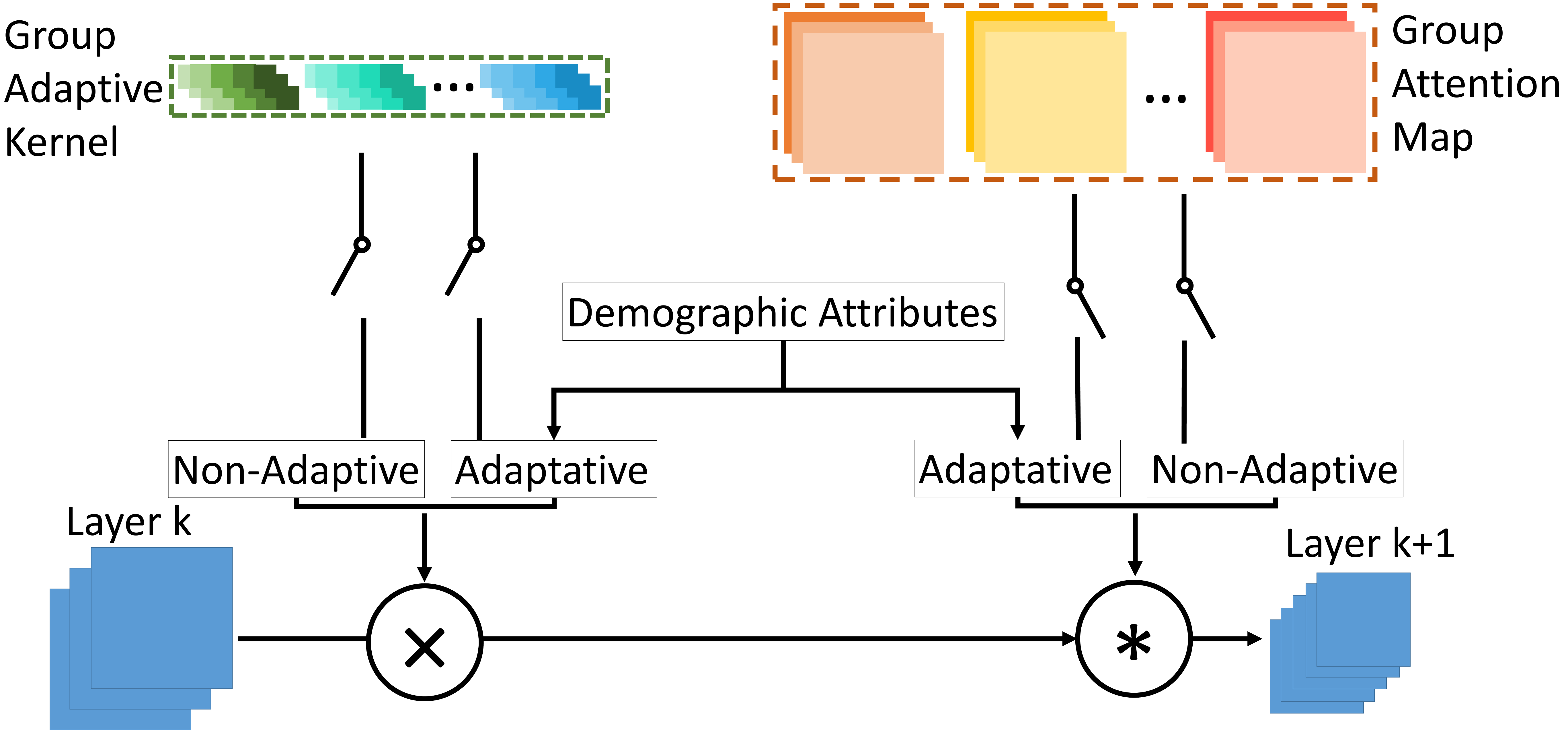}
	    \caption{{\footnotesize GAC}}
	    \label{fig:gac}
	    \end{subfigure}\\	  
	    \vspace{-3mm}
	    \caption{\footnotesize{A comparison of approaches in adaptive CNNs.}}
	    \label{fig:related_work}
\end{figure*}

\textbf{Adaptive Neural Networks.} Three types of CNN-based adaptive learning techniques are related to our work: adaptive architectures, adaptive kernels, and attention mechanism. 
Adaptive architectures design new performance-based neural functions or structures, {\it e.g.}, neuron selection hidden layers~\cite{hu2018learning} and automatic CNN expansion for FR~\cite{zhang2016adaptive}. 
As CNN advances many fields, prior works propose dynamic kernels to realize content-adaptive convolutions. 
Li~\etal~\cite{li2015shape} propose a shape-driven kernel for facial trait recognition where each  landmark-centered patch has a unique kernel. 
A convolution fusion for graph neural networks is introduced by~\cite{du2017topology} where  varying-size filters are used per layer. 
The works of~\cite{ding2018automatic,li2019selective} use a kernel selection scheme to automatically adjust the receptive field size based on inputs. 
To better suit input data, ~\cite{ding2017convolutional} splits training data into clusters and learns an exclusive kernel per cluster. 
Li~\etal~\cite{li2017adaptive} introduce an adaptive CNN for object detection that transfers pre-trained CNNs to a target domain by selecting useful kernels per layer. 
Alternatively, one may feed input images  into a  kernel function to dynamically generate kernels~\cite{su2019pixel, zamora2019adaptive, klein2015dynamic, jia2016dynamic}. 
Despite its effectiveness, such individual adaptation may not be suitable given the diversity of faces in demographic groups. 
Our work is most related to the side information adaptive convolution~\cite{kang2017incorporating}, where in each layer a sub-network inputs auxiliary information to generate filter weights.
We mainly differ in that GAC automatically learns where to use adaptive kernels in a multi-layer CNN (see Figs.~\ref{fig:adap_kernel} and~\ref{fig:gac}), thus more efficient and capable in applying to a deeper CNN.

As the human perception naturally selects the most pertinent piece of information, attention mechanisms are designed for many tasks, {\it e.g.}, detection~\cite{zhang2018progressive}, recognition~\cite{chen2018learning}, image captioning~\cite{chen2017sca},  tracking~\cite{chen2019multi}, pose estimation~\cite{su2019multi}, and segmentation~\cite{lu2019see}. 
Normally, attention weights are estimated by feeding images or feature maps into a shared network, composed of convolutional and pooling layers~\cite{bastidas2019channel, chen2018learning, ling2020attention, sindagi2019ha} or multi-layer perceptron (MLP)~\cite{hu2018squeeze, woo2018cbam, sadiq2019facial, linsley2019learning}. 
Apart from feature-based attention, Hou~\etal~\cite{hou2019cross} propose a correlation-guided cross attention map for few-shot classification where the correlation between the class feature and query feature generates attention weights. 
The work of~\cite{yang2019cross} introduces a cross-channel communication block to encourage information exchange across  channels. 
To accelerate channel interaction, Wang~\etal~\cite{wang2019eca} propose a $1$D convolution across channels for attention prediction. 
Different from prior work, our attention maps are constructed by demographic information (see Figs.~\ref{fig:attention_map},~\ref{fig:gac}), which improves the robustness of face representations in every demographic group.

\Section{Methodology}
\SubSection{Overview}
Our goal is to train a FR network that is impartial to individuals in different demographic groups. 
Unlike image-related variations, {\it e.g.}, large-poses or low-resolution faces are harder to be recognized, demographic attributes are subject-related properties with no apparent impact in recognizability of identity, at least from a layman's perspective. 
Thus, an unbiased FR system should be able to obtain equally salient features for faces across demographic groups. 
However, due to imbalanced demographic distributions and inherent face differences between groups, it was shown that  certain groups achieve higher performance even with hand-crafted features~\cite{klare2012face}. 
Thus, it is impractical to extract features from different demographic groups that exhibit equal discriminability. 
Despite such disparity, a FR algorithm can still be designed to {\it mitigate} the difference in performance. 

\begin{figure*}[t!]
\captionsetup{font=footnotesize}
    \centering
    \includegraphics[width=0.96\linewidth]{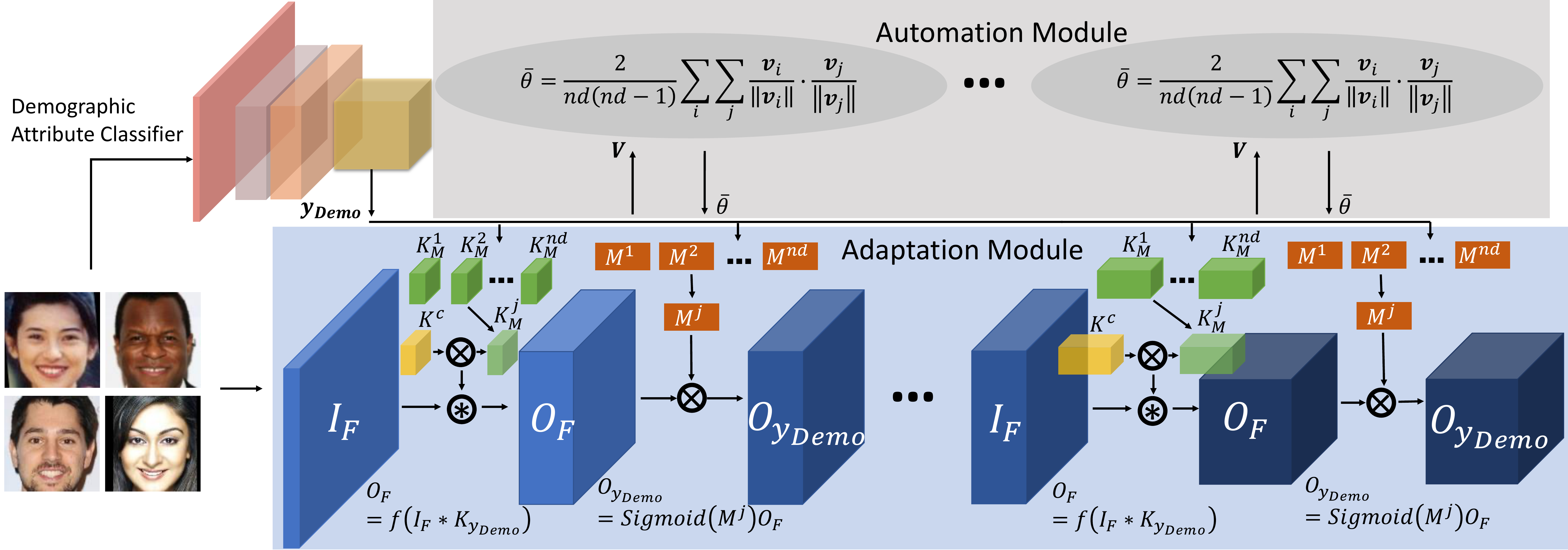}
    \caption{\footnotesize{Overview of the proposed GAC for mitigating FR bias. GAC contains two major modules: the adaptive layer and the automation module. The adaptive layer consists of adaptive kernels and attention maps. The automation module is employed to decide whether a layer should be adaptive or not.}} 
    \label{fig:overview} \figvspace
\end{figure*}

To this end, we propose a CNN-based group adaptive classifier to utilize dynamic kernels and attention maps to boost FR performance in all demographic groups considered here.
Specifically, GAC has two main modules, an adaptive layer and an automation module. In adaptive layer, face images or feature maps are convolved with a unique kernel for each demographic group, and multiplied with adaptive attention maps to obtain demographic-differential features for faces in a certain group. The automation module determines in which layers of the network adaptive kernels and attention maps should be applied. As shown in Fig.~\ref{fig:overview}, given an aligned face, and its identity label $y_{ID}$, a pre-trained demographic classifier first estimates its demographic attribute $y_{Demo}$. With $y_{Demo}$, the image is then fed into a recognition network with multiple demographic adaptive layers to estimate its identity. In the following, we present these two modules.

\SubSection{Adaptive Layer}
\textit{Adaptive Convolution.} For a standard convolution operation in CNN, an image or feature map from the previous layer $I_{F} \in \mathbb{R}^{ic \times ih \times iw}$ is convolved with a single kernel matrix $K \in \mathbb{R}^{kc \times ic \times kh \times kw}$, where $ic$ is the number of input channels, $kc$ the number of filters, $ih$ and $iw$ the input size, and $kh$ and $kw$ the filter size. Such an operation shares the kernel with every input that goes through the layer, and is thus agnostic to demographic content, resulting in limited capacity to represent faces of minority groups. 
To mitigate the bias in convolution, we introduce a trainable matrix of kernel masks $K_{M} \in \mathbb{R}^{nd \times ic \times kh \times kw}$, where $nd$ is the number of demographic groups. During the forward pass, the demographic label $y_{Demo}$ and kernel matrix $K_{M}$ are fed into the adaptive convolutional layer to generate demographic adaptive filters.
Let $K^c \in \mathbb{R}^{ic \times kh \times kw}$ denote the $c^{th}$ channel filter, and the adaptive filter weights for $c^{th}$ channel are:
\begin{equation}
    K_{y_{Demo}}^{c} = K^c \mbox{\small $\bigotimes$} K^{j}_{M},
\end{equation}
where $K^{j}_{M} \in \mathbb{R}^{ic \times kh \times kw}$ is the ${j}^{th}$ kernel mask for group $y_{Demo}$, and $\bigotimes$ denotes element-wise multiplication. Then the $c^{th}$ channel of the output feature map is given by $O_F^{c} = f(I_{F} * K_{y_{Demo}}^{c})$, where * denotes convolution, and $f(\cdot)$ is  activation. Differ to the conventional convolution, samples in every demographic group have a unique kernel $K_{y_{Demo}}$.

\textit{Adaptive Attention.} Each channel filter in a CNN plays an important role in every dimension of the final representation, which can be viewed as a semantic pattern detector~\cite{chen2017sca}. In the adaptive convolution, however, the values of a kernel mask are broadcast along the channel dimension, indicating that the weight selection is spatially varied but channel-wise joint. Hence, we introduce a channel-wise attention mechanism to enhance the face features that are demographic-adaptive. First, a trainable matrix of channel attention maps $M \in \mathbb{R}^{nd \times kc}$ is initialized in every adaptive attention layer. Given $y_{Demo}$ and the current feature map $O_F \in \mathbb{R}^{kc \times oh \times ow}$, where $oh$ and $ow$ are the height and width of $O_F$, the $c^{th}$ channel of the new feature map is calculated by:
\begin{equation}
    O^{c}_{y_{Demo}} = \textrm{Sigmoid}({M}^{jc}) \cdot O_F^c,
\end{equation}
where ${M}^{jc}$ is the entry in the $j^{th}$ row of $M$ for the demographic group $y_{Demo}$ at $c^{th}$ column.
In contrast to the adaptive convolution, elements of each demographic attention map ${M}^{j}$ diverge in channel-wise manner, while the single attention weight ${M}^{jc}$ is spatially shared by the entire matrix $O_F^c \in \mathbb{R}^{oh \times ow}$. The two adaptive matrices, $K_M$ and
${M}$, are jointly tuned with all the other parameters supervised by the classification loss.

Unlike dynamic CNNs~\cite{kang2017incorporating} where additional networks are engaged to produce input-variant kernel or attention map, our adaptiveness is yielded by a simple thresholding function directly pointing to the demographic group with no auxiliary networks. Although the kernel network in~\cite{kang2017incorporating} can generate continuous kernels without enlarging the parameter space, further encoding is required if the side inputs for kernel network are discrete variables. Our approach, in contrast, divides kernels into clusters so that the branch parameter learning can stick to a specific group without interference from individual uncertainties, making it suitable for discrete domain adaptation. Further, the adaptive kernel masks in GAC are more efficient in terms of the number of additional parameters. Compared to a non-adaptive layer, the number of additional parameters of GAC is $nd \times ic \times kh \times kw$, while that of ~\cite{kang2017incorporating} is $id \times kc \times ic \times kh \times kw$ if the kernel network is a one-layer MLP,
where $id$ is the dimension of input side information.
Thus, for one adaptive layer, ~\cite{kang2017incorporating} has $\frac{id\times kc}{nd}$ times more parameters than ours, which can be substantial given the typical large value of $kc$, the number of filters.

\SubSection{Automation Module}
Though faces in different demographic groups are adaptively processed by various kernels and attention maps, it is inefficient to use such adaptations in {\it every} layer of a deep CNN. 
To relieve the burden of unnecessary parameters and avoid empirical trimming, we adopt a similarity fusion process to automatically determine the adaptive layers. 
Since the same fusion scheme can be applied to both types of adaptation, we take the adaptive convolution as an example to illustrate this automatic scheme. 

First, a matrix composed of $nd$ kernel masks is initialized in every convolutional layer. As training continues, each kernel mask is updated independently to reduce classification loss for each demographic group. Second, we reshape the kernel masks into $1$D vectors $\mathbf{V} = [\mathbf{v}_1, \mathbf{v}_2, \ldots, \mathbf{v}_{nd}]$, where $\mathbf{v}_i \in \mathbb{R}^{l}, l=ic \times kw \times kh$ is the kernel mask of the $i^{th}$ demographic group. Next, we compute Cosine similarity between two kernel vectors,
    $\theta_{ij} = \frac{\mathbf{v}_i}{\|\mathbf{v}_i\|} \cdot \frac{\mathbf{v}_j}{\|\mathbf{v}_j\|}$,
where $1\leq i,j\leq nd$. The average similarity of all pair-wise similarities is obtained by $\overline{\theta} = \frac{2}{nd(nd-1)}\sum_i \sum_j \theta_{ij}, i\neq j$. If $\overline{\theta}$ is higher than a pre-defined threshold $\tau$, the kernel parameters in this layer reveal the demographic-agnostic property. Hence, we merge the $nd$ kernels into a single kernel by taking the average along the group dimension. In the subsequent training, this single kernel can still be updated separately for each demographic group, since the kernel may become demographic-adaptive in later epochs. We monitor the similarity trend of the adaptive kernels in each layer until $\overline{\theta}$ is stable.


\SubSection{De-biasing Objective Function}
Apart from the objective function for face identity classification, we also adopt a regress loss function to narrow the gap of the intra-class distance between demographic groups. Let $g(\cdot)$ denote the inference function of GAC, and $I^{i}_{jg}$ is the $i^{th}$ image of subject $j$ in group $g$. Thus, the feature representation of image $I^{i}_{jg}$ is given by $\mathbf{r}_{jg}^i = g(I^{i}_{jg}, \mathbf{w}_r)$, where $\mathbf{w}_r$ denotes the GAC parameters. 
Assuming the feature distribution of each subject is a Gaussian distribution with diagonal covariance matrix (axis-aligned hyper-ellipsoid), we utilize mahalanobis distance as the intra-class distance of each subject. In particular, we first compute the center point of each identity-ellipsoid:
\begin{equation}
    \boldsymbol{\mu}_{jg} = \frac{1}{N} \sum_{i=1}^N g(I^{i}_{jg}, \mathbf{w}_r),
\end{equation}
where $N$ is the total number of face images of subject $j$. The average intra-class distance of subject $j$ is as follows: 
\begin{equation}
    Dist_{jg} = \frac{1}{N} \sum_{i=1}^N (\mathbf{r}_{jg}^i - \boldsymbol{\mu}_{jg})^T (\mathbf{r}_{jg}^i - \boldsymbol{\mu}_{jg}).
\end{equation}
We then compute the intra-class distance for all subjects in group $g$ as $Dist_{g} = \frac{1}{Q} \sum_{j=1}^{Q} Dist_{jg}$, where $Q$ is the number of total subjects in group $g$. This allows us to lower the difference of intra-class distance by:
\begin{equation}
    \mathcal{L}_{bias} = \frac{\lambda}{Q \times nd} \sum_{j=1} ^ {Q \times nd} \Bigl | Dist_{jg} - \frac{1}{nd} \sum_{g=1} ^ {nd} Dist_{g} \Bigr |,
\end{equation}
where $\lambda$ is the coefficient for the de-biasing objective.

\Section{Experiments}
\textbf{Datasets}: Our bias study uses RFW dataset~\cite{wang2019racial} for testing and BUPT-Balancedface dataset~\cite{wang2020mitigating} for training. 
RFW consists of faces in four race/ethnic groups: White, Black, East Asian, and South Asian~\footnote{ RFW~\cite{wang2019racial} uses Caucasian, African, Asian, and Indian to name demographic groups. We adopt these groups and accordingly rename to White, Black, East Asian, and South Asian for clearer race/ethnicity definition.}.
Each group contains $\sim$$10$K images of $3$K individuals for face verification. BUPT-Balancedface contains $1.3$M images of $28$K celebrities and is approximately race-balanced with $7$K identities per race. 
Other than race, we also study gender bias. 
We combine IMDB~\cite{rothe2018deep}, UTKFace~\cite{zhifei2017cvpr}, AgeDB~\cite{moschoglou2017agedb}, AAF~\cite{cheng2019exploiting}, AFAD~\cite{niu2016ordinal} to train a gender classifier, which estimates gender of faces in RFW and BUPT-Balancedface. 
All face images are cropped and resized to $112 \times 112$ pixels via landmarks detected by RetinaFace~\cite{deng2019retinaface}.

\textbf{Implementation Details}: We train a baseline network and GAC on BUPT-Balancedface, using the $50$-layer ArcFace architecture~\cite{Deng_2019_CVPR}.
The classification loss is an additive Cosine margin in Cosface~\cite{wang2018cosface}, with the scale and margin of $s=64$ and $m=0.5$. Training is optimized by  and a batch size $256$. The learning rate starts from $0.1$ and drops to $0.0001$ following the schedule at $8$, $13$, $15$ epochs for the baseline, and $5$, $17$, $19$ epochs for GAC. 
We set $\lambda=0.1$ for the intra-distance de-biasing. $\tau=-0.2$ is chosen for automatic adaptation in GAC. Our FR models are trained to extract a  $512$-dim representation. Our demographic classifier uses a $18$-layer ResNet~\cite{he2016deep}.  Comparing the GAC and baseline, the average feature extraction speed per image on Nvidia $1080$Ti GPU is $1.4$ms and $1.1$ms,  and the number of model parameters is $44.0$M and $43.6$M, respectively.

\textbf{Performance Metrics}:
The common group fairness criteria like demographic parity distance are improper to evaluate fairness of learnt representations, since they are typically designed to measure independence properties of random variables.
However, in FR the sensitive demographic characteristics are tied to identities, making these two variables correlated. 
The NIST report proposes to use false negative and false positive for each demographic group to measure the fairness~\cite{grother2019frvt}.
Instead of plotting false negative vs.~false positives, we adopt a compact quantitative metric, {\it i.e.}, the standard deviation (STD) of the performance in different demographic groups, that was previously introduced in~\cite{wang2020mitigating,gong2020jointly} and called ``biasness''.
We also report average accuracy (Avg) to show the overall FR performance.

\SubSection{Results on RFW Protocol}
We follow RFW face verification protocol with $6$K pairs per race/ethnicity. 
The models are trained on BUPT-Balancedface with ground truth race and identity labels. 

\textbf{Compare with SOTA.} We compare the GAC with four SOTA algorithms on RFW protocol, namely, ACNN~\cite{kang2017incorporating}, RL-RBN~\cite{wang2020mitigating}, PFE~\cite{shi2019probabilistic}, and DebFace~\cite{gong2020jointly}. Since the approach in ACNN~\cite{kang2017incorporating} is related to GAC, we re-implement it and apply to the  bias mitigation problem. First, we train a race classifier with the cross-entropy loss on BUPT-Balancedface. Then the softmax output of our race classifier is fed to a filter manifold network (FMN) to generate adaptive filter weights. Here, FMN is a two-layer MLP with a ReLU in between. Similar to GAC, race probabilities are considered as auxiliary information for face representation learning. We also compare with the SOTA approach PFE~\cite{shi2019probabilistic} by training it on BUPT-Balancedface. As shown in Tab.~\ref{tab:bias_rfw_sota}, 
GAC is superior to SOTA w.r.t.~average performance and feature fairness. 
Compared to kernel masks in GAC, the FMN in ACNN~\cite{kang2017incorporating} contains more trainable parameters. Applying it to each convolutional layer is prone to overfitting. In fact, eight layers are empirically chosen as the FMN based convolution.
As the race data is a four-element input in our case, using extra kernel networks adds complexity to the FR network, which degrades the verification performance.
Even though PFE performs the best on standard benchmarks (Tab.~\ref{tab:lfw_ijba_ijbc}), it still exhibits high biasness. Our GAC outperforms PFE on RFW in both biasness and average performance. Compared to DebFace~\cite{gong2020jointly}, in which demographic attributes are disentangled from the identity representations, GAC achieves higher verification performance by optimizing the classification for each demographic group, with a lower biasness as well.

\begin{table}[t]
    \captionsetup{font=footnotesize}
    \centering
    \caption{\footnotesize Performance comparison with SOTA on the RFW protocol~\cite{wang2019racial}. The results marked by (*) are directly copied from~\cite{wang2020mitigating}.}
    \label{tab:bias_rfw_sota}
   \scalebox{0.69}{
    \begin{tabular}{@{}c cccc cc@{}}
        \toprule
        Method & White & Black & East Asian & South Asian & Avg ($\uparrow$) & STD ($\downarrow$)\\
        \midrule
        RL-RBN~\cite{wang2020mitigating} & $96.27$ & $95.00$ & $94.82$ & $94.68$ & $95.19$ & $0.63$ \\
        ACNN~\cite{kang2017incorporating} & $96.12$ & $94.00$ & $93.67$ & $94.55$ & $94.58$ & $0.94$\\
        PFE~\cite{shi2019probabilistic} & $\mathbf{96.38}$ & $\mathbf{95.17}$ & $94.27$ & $94.60$ & $95.11$ & $0.93$\\
        ArcFace~\cite{Deng_2019_CVPR} & $96.18^*$ & $94.67^*$ & $93.72^*$ & $93.98^*$ & $94.64$ & $0.96$\\
        CosFace~\cite{wang2018cosface} & $95.12^*$ & $93.93^*$ & $92.98^*$ & $92.93^*$ & $93.74$ & $0.89$\\
        DebFace~\cite{gong2020jointly} & $95.95$ & $93.67$ & $94.33$ & $94.78$ & $94.68$ & $0.83$\\
        GAC & $96.20$ & $94.77$ & $\mathbf{94.87}$ & $\mathbf{94.98}$ & $\mathbf{95.21}$ & $\mathbf{0.58}$\\
        \bottomrule
    \end{tabular}}
\end{table}

\textbf{Ablation.} To investigate the efficacy of adaptive layers, automation module, and de-biasing loss in GAC, we conduct three ablation studies: adaptive mechanisms, number of convolutional layers, and demographic information. For adaptive mechanisms, since deep feature maps contain both spatial and channel-wise information, we study the relationship among adaptive kernels, spatial and channel-wise attentions, and their impact to bias mitigation. We also study the impact of $\tau$ in our automation module. Apart from the baseline and GAC, we ablate eight variants: (1) GAC-Channel: channel-wise attention for race-differential feature; (2) GAC-Kernel: adaptive convolution with race-specific kernels; (3) GAC-Spatial: only spatial attention is added to baseline; (4) GAC-CS: both channel-wise and spatial attention; (5) GAC-CSK: combine adaptive convolution with spatial and channel-wise attention; (6,7,8) GAC-($\tau=*$): set $\tau$ to $*$.

\begin{table}[t]
    \captionsetup{font=footnotesize}
    \centering
    \caption{\footnotesize  Ablation of adaptive mechanisms on the RFW protocol~\cite{wang2019racial}.}
    \label{tab:bias_rfw_adap}
   \scalebox{0.67}{
    \begin{tabular}{@{}c cccc cc@{}}
        \toprule
        Method & White & Black & East Asian & South Asian &  Avg ($\uparrow$) & STD ($\downarrow$)\\
        \midrule
        Baseline & $96.18$ & $93.98$ & $93.72$ & $94.67$ & $94.64$ & $1.11$ \\
        GAC-Channel & $95.95$ & $93.67$ & $94.33$ & $94.78$ & $94.68$ & $0.83$ \\
        GAC-Kernel & $96.23$ & $94.40$ & $94.27$ & $94.80$ & $94.93$ & $0.78$ \\
        GAC-Spatial & $95.97$ & $93.20$ & $93.67$ & $93.93$ & $94.19$ & $1.06$ \\
        GAC-CS & $96.22$ & $93.95$ & $94.32$ & $\mathbf{95.12}$ & $94.65$ & $0.87$ \\
        GAC-CSK & $96.18$ & $93.58$ & $94.28$ & $94.83$ & $94.72$ & $0.95$ \\
        GAC-($\tau=0$) & $96.18$ & $93.97$ & $93.88$ & $94.77$ & $94.70$ & $0.92$ \\
        GAC-($\tau=-0.1$) & $\mathbf{96.25}$ & $94.25$ & $94.83$ & $94.72$ & $95.01$ & $0.75$ \\
        GAC-($\tau=-0.2$) & $96.20$ & $\mathbf{94.77}$ & $\mathbf{94.87}$ & $94.98$ & $\mathbf{95.21}$ & $\mathbf{0.58}$\\
        \bottomrule
    \end{tabular}}
\end{table}

\begin{figure}[t]
\centering
\captionsetup{font=footnotesize}
    \includegraphics[width=0.7\linewidth]{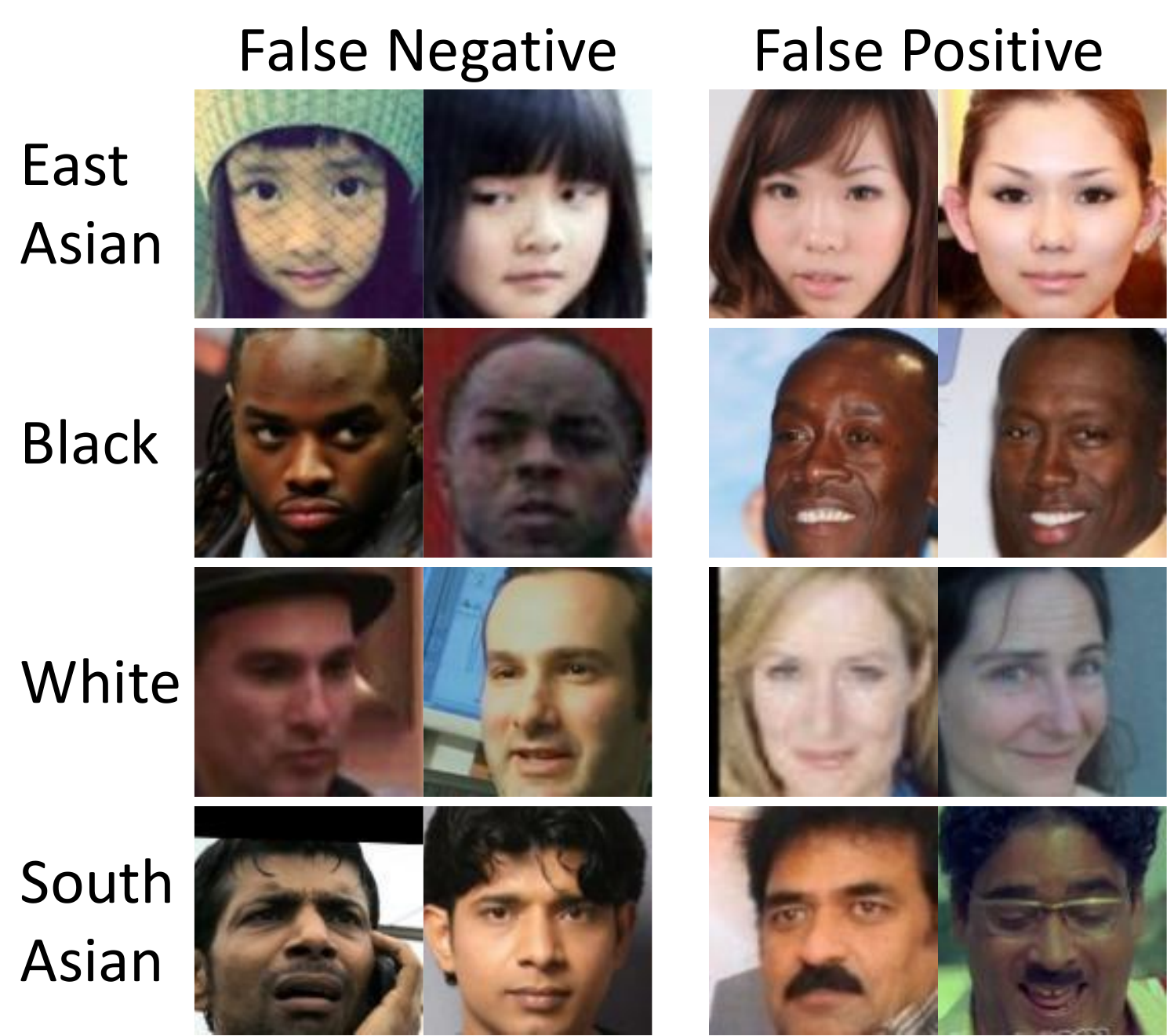}
    \captionof{figure}{\footnotesize{$8$ false positive and false negative pairs on RFW given by the baseline but successfully verified by GAC.}}
    \label{fig:failexamples}
\end{figure}

Given the ablation results in Tab.~\ref{tab:bias_rfw_adap}, 
we make several observations: (1) the baseline model is the most biased across race groups. 
(2) spatial attention mitigates the race bias at the cost of verification accuracy, and is less effective on learning fair features than other adaptive techniques. This is probably because spatial contents, especially local layout information, only reside at earlier CNN layers, where the spatial dimensions are gradually decreased by the later convolutions and poolings. 
Thus, semantic details like demographic attributes are hardly encoded spatially. 
(3) Compared to GAC, combining adaptive kernels with both spatial and channel-wise attention increases the number of parameters, lowering the performance. 
(4) As $\tau$ determines the number of adaptive layers in GAC, it has a great impact on the performance. A small $\tau$ may increase redundant adaptive layers, while the adaptation layers may lack in capacity if too large.
Fig.~\ref{fig:failexamples} shows pairs of false positives (two faces falsely verified as the same identity) and false negatives
(two faces falsely verified as different identities) produced by the baseline but successfully verified by GAC.

\begin{table}[t]
    \captionsetup{font=footnotesize}
    \centering
    \caption{\footnotesize Ablation of CNN depths and demographics on RFW protocol~\cite{wang2019racial}.}
    \label{tab:bias_rfw_depth}
   \scalebox{0.67}{
    \begin{tabular}{@{}c cccc cc@{}}
        \toprule
        Method & White & Black & East Asian & South Asian &  Avg ($\uparrow$) & STD ($\downarrow$)\\
        \midrule
        \multicolumn{7}{c}{Number of Layers}\\
        ArcFace-34 & $96.13$ & $93.15$ & $92.85$ & $93.03$ & $93.78$ & $1.36$ \\
        GAC-ArcFace-34 & $96.02$ & $94.12$ & $94.10$ & $94.22$ & $94.62$ & $0.81$ \\
        ArcFace-50 & $96.18$ & $93.98$ & $93.72$ & $94.67$ & $94.64$ & $1.11$ \\
        GAC-ArcFace-50 & $96.20$ & $94.77$ & $94.87$ & $94.98$ & $95.21$ & $0.58$\\
        ArcFace-100 & $96.23$ & $93.83$ & $94.27$ & $94.80$ & $94.78$ & $0.91$ \\
        GAC-ArcFace-100 & $96.43$ & $94.53$ & $94.90$ & $95.03$ & $95.22$ & $0.72$ \\
        \midrule
        \multicolumn{7}{c}{Race/Ethnicity Labels}\\
        Ground-truth & $96.20$ & $94.77$ & $94.87$ & $94.98$ & $95.21$ & $0.58$\\
        Estimated & $96.27$ & $94.40$ & $94.32$ & $94.77$ & $94.94$ & $0.79$ \\
        Random & $95.95$ & $93.10$ & $94.18$ & $94.82$ & $94.50$ & $1.03$ \\
        \bottomrule
    \end{tabular}}
\end{table}

\begin{table*}[t!]
    \captionsetup{font=footnotesize}
    \centering
    \scalebox{0.75}{
    \begin{tabular}{c cccccc c}
        \toprule
        Method & Gender & White & Black & East Asian & South Asian & Avg ($\uparrow$) & STD ($\downarrow$)\\
        \midrule
        \multirow{2}{*}{Baseline} & Male & $97.49 \pm 0.08$ & $96.94 \pm 0.26$ & $97.29 \pm 0.09$ & $97.03 \pm 0.13$ & \multirow{2}{*}{$96.96 \pm 0.03$} & \multirow{2}{*}{$0.69 \pm 0.04$}\\
        & Female & $97.19 \pm 0.10$ & $97.93 \pm 0.11$ & $95.71 \pm 0.11$ & $96.01 \pm 0.08$ & &\\
        \multirow{2}{*}{AL+Manual} & Male & $98.57 \pm 0.10$ & $98.05 \pm 0.17$ & $98.50 \pm 0.12$ & $\mathbf{98.36 \pm 0.02}$ & \multirow{2}{*}{$98.09 \pm 0.05$} & \multirow{2}{*}{$0.66 \pm 0.07$}\\
        & Female & $98.12 \pm 0.18$ & $\mathbf{98.97 \pm 0.13}$ & $96.83 \pm 0.19$ & $97.33 \pm 0.13$ & &\\
        \multirow{2}{*}{GAC} & Male & $\mathbf{98.75 \pm 0.04}$ & $\mathbf{98.18 \pm 0.20}$ & $\mathbf{98.55 \pm 0.07}$ & $98.31 \pm 0.12$ & \multirow{2}{*}{$\mathbf{98.19 \pm 0.06}$} & \multirow{2}{*}{$\mathbf{0.56 \pm 0.05}$}\\
        & Female & $\mathbf{98.26 \pm 0.16}$ & $98.80 \pm 0.15$ & $\mathbf{97.09 \pm 0.12}$ & $\mathbf{97.56 \pm 0.10}$ & &\\
        \bottomrule
    \end{tabular}}
    \tablespace
        \caption{\footnotesize Verification Accuracy (\%) of $5$-fold cross-validation on $8$ groups of RFW~\cite{wang2019racial}.}
        \label{tab:bias_gender_race}
\end{table*}

Both the adaptive layers and de-biasing loss in GAC can be applied to CNN in any depth. In this ablation, we train both the baseline and GAC in ArcFace architecture with three different numbers of layers): (1) ArcFace-34/GAC-ArcFace-34: 34-layer CNN, (2) ArcFace-50/GAC-ArcFace-50, and (3) ArcFace-100/GAC-ArcFace-100. 
As the training of GAC relies on demographic information, the error and bias in demographic labels might influence the FR bias reduction of GAC. Thus, we conduct three experiments with different demographic information, (1) ground-truth: the race/ethnicity labels provided by RFW; (2) estimated: the labels predicted by a pre-trained race estimation model; (3) random: the demographic label randomly assigned to each face image.

Tab.~\ref{tab:bias_rfw_depth} reports the ablation results of depths and demographic information. Compared to the baseline models, GAC reduces the performance STD for all the backbones with different number of layers. We see that the model with least number of layers presents the most bias, and the reduction of biasness by GAC is the most as well. The noise in demographic labels does, however, impair the performance of GAC. With estimated demographic information, the biasness is higher than that of the model with ground-truth supervision. Meanwhile, the model trained with randomly assigned demographics has the highest biasness.

\SubSection{Results on Gender and Race Groups}
\label{sec:gender_race}
We now extend demographic attributes to both gender and race. First, we train two classifiers that predict gender and race/ethnicity of a face image. The classification accuracy of gender and race/ethnicity is $85\%$ and $81\%$\footnote{This seemingly low accuracy is mainly due to the large dataset we assembled for training and testing gender/race classifiers. Our demographic classifier has been shown to perform comparably as SOTA on common benchmarks.  While demographic estimation errors impact the training, testing, and evaluation of bias mitigation algorithms, the evaluation is of the most concern as demographic label errors  may greatly impact the biasness calculation. Thus, future development may include  either manually cleaning the labels, or designing a biasness metric robust to label errors.}, respectively. Then, these fixed classifiers are affiliated with GAC to provide demographic information for learning adaptive kernels and attention maps. We merge BUPT-Balancedface and RFW, and split the subjects into $5$ sets for each of $8$ demographic groups. In $5$-fold cross-validation, each time a model is trained on $4$ sets and tested on the remaining set.

Here we demonstrate the efficacy of the automation module for GAC. 
We compare to the scheme of manually design (AL+Manual) that adds adaptive kernels and attention maps to a subset of layers. 
Specifically, the first block in every residual unit is chosen to be the adaptive convolution layer, and channel-wise attentions are applied to the feature map output by the last block in each residual unit. 
As we use $4$ residual units and each block has $2$ convolutional layers, the manual scheme involves $8$ adaptive convolutional layers and $4$ groups of channel-wise attention maps.
As in Tab.~\ref{tab:bias_gender_race}, automatic adaptation is more effective in enhancing the discirminability and fairness of face representations. 
Figure~\ref{fig:kernel_step} shows the dissimilarity of kernel masks in the convolutional layers changes during training epochs under three thresholds $\tau$. A lower $\tau$ results in more adaptive layers.
We see the layers that are determined to be adaptive do vary across both layers (vertically) and training time (horizontally), which shows the importance of our automatic mechanism.

\begin{figure*}[t!]

	    \captionsetup{font=footnotesize}
	    \captionsetup[subfigure]{labelformat=empty}
	    \centering
	    \vspace{-3mm}
	    \begin{subfigure}[b]{0.8\linewidth}
	    \centering
	    \begin{minipage}[c]{0.06\textwidth}
	    \caption{{\tiny Average Image}}
	    \end{minipage}\hfill
	    \begin{minipage}[c]{0.94\textwidth}
	    \begin{subfigure}[b]{0.125\linewidth}
	    \caption{{\tiny East Asian female}}
		\includegraphics[width=\linewidth]{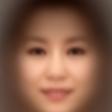}
		\end{subfigure}\hfill
		\begin{subfigure}[b]{0.125\linewidth}
		\caption{{\tiny East Asian male}}
		\includegraphics[width=\linewidth]{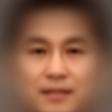}
		\end{subfigure}\hfill
		\begin{subfigure}[b]{0.125\linewidth}
		\caption{{\tiny Black female}}
		\includegraphics[width=\linewidth]{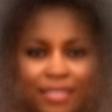}
		\end{subfigure}\hfill
		\begin{subfigure}[b]{0.125\linewidth}
		\caption{{\tiny Black male}}
		\includegraphics[width=\linewidth]{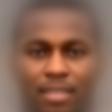}
		\end{subfigure}\hfill
		\begin{subfigure}[b]{0.125\linewidth}
		\caption{{\tiny White female}}
		\includegraphics[width=\linewidth]{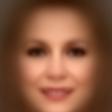}
		\end{subfigure}\hfill
		\begin{subfigure}[b]{0.125\linewidth}
		\caption{{\tiny White male}}
		\includegraphics[width=\linewidth]{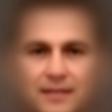}
		\end{subfigure}\hfill
		\begin{subfigure}[b]{0.125\linewidth}
		\caption{{\tiny South Asian female}}
		\includegraphics[width=\linewidth]{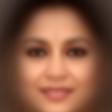}
		\end{subfigure}\hfill
		\begin{subfigure}[b]{0.125\linewidth}
		\caption{{\tiny South Asian male}}
		\includegraphics[width=\linewidth]{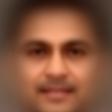}
	    \end{subfigure}
	    \end{minipage}
	    \end{subfigure}
	     
	    \vspace{3px}
	    \begin{subfigure}[b]{0.8\linewidth}
		\centering
		\begin{minipage}[c]{0.06\textwidth}
	    \caption{{\tiny Heatmap +Image -GAC}}
	    \end{minipage}\hfill
	    \begin{minipage}[c]{0.94\textwidth}
	    \includegraphics[width=0.125\linewidth]{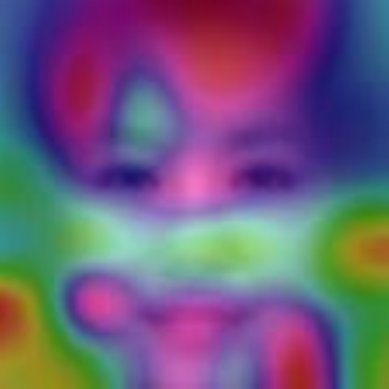}\hfill
	    \includegraphics[width=0.125\linewidth]{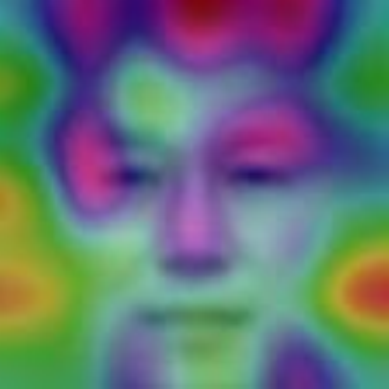}\hfill
	    \includegraphics[width=0.125\linewidth]{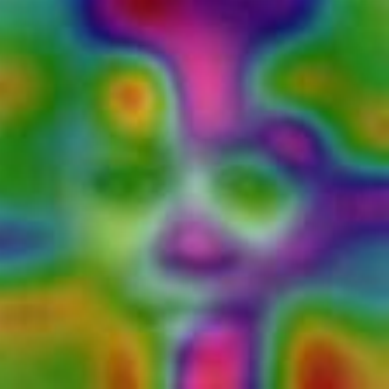}\hfill
	    \includegraphics[width=0.125\linewidth]{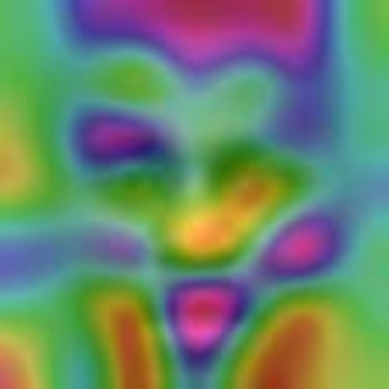}\hfill
	    \includegraphics[width=0.125\linewidth]{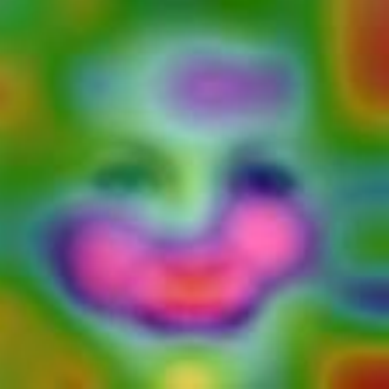}\hfill
	    \includegraphics[width=0.125\linewidth]{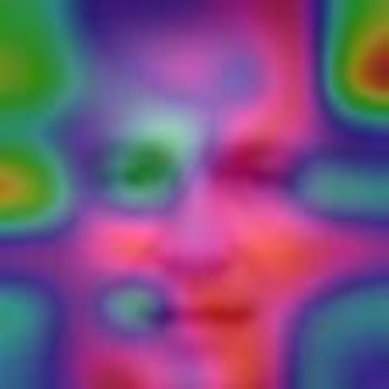}\hfill
	    \includegraphics[width=0.125\linewidth]{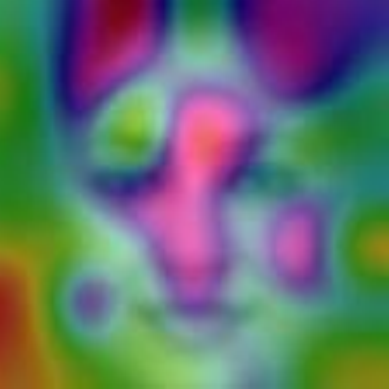}\hfill
	    \includegraphics[width=0.125\linewidth]{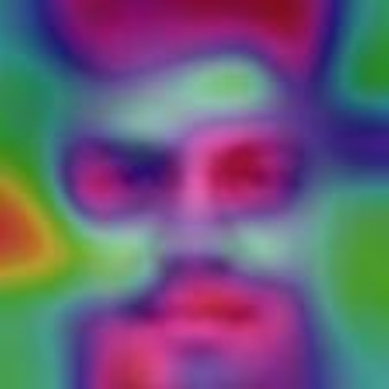}
	    \end{minipage}
	     \end{subfigure}\\
	    \begin{subfigure}[b]{0.8\linewidth}
		\centering
		\begin{minipage}[c]{0.06\textwidth}
	    \caption{{\tiny Heatmap +Image -Base}}
	    \end{minipage}\hfill
	    \begin{minipage}[c]{0.94\textwidth}
	    \includegraphics[width=0.125\linewidth]{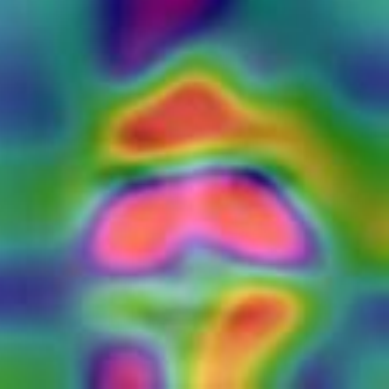}\hfill
	    \includegraphics[width=0.125\linewidth]{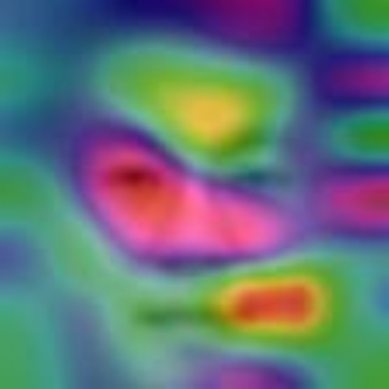}\hfill
	    \includegraphics[width=0.125\linewidth]{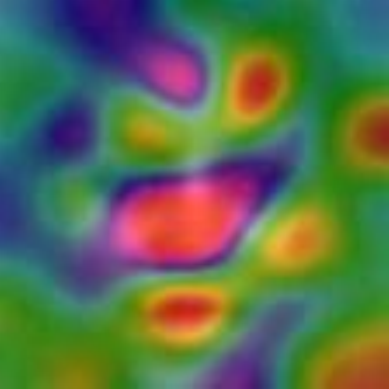}\hfill
	    \includegraphics[width=0.125\linewidth]{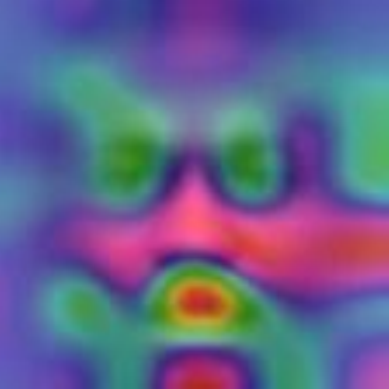}\hfill
	    \includegraphics[width=0.125\linewidth]{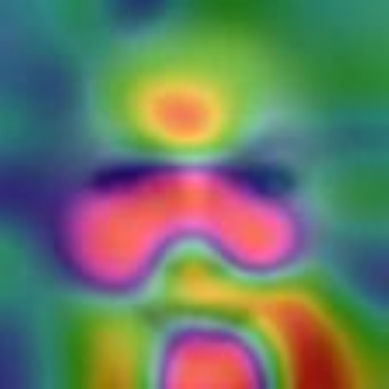}\hfill
	    \includegraphics[width=0.125\linewidth]{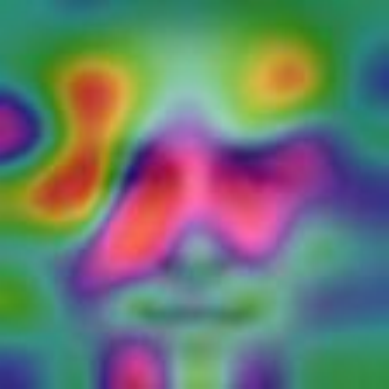}\hfill
	    \includegraphics[width=0.125\linewidth]{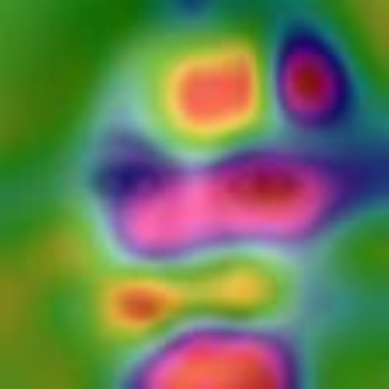}\hfill
	    \includegraphics[width=0.125\linewidth]{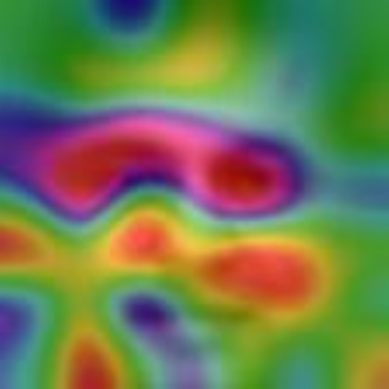}
	    \end{minipage}
	     \end{subfigure}

	    \caption{\footnotesize The first row shows the average faces of different groups in RFW. The next two rows show  gradient-weighted class activation heatmaps~\cite{selvaraju2017grad} at the $43^{th}$ convolutional layer of the GAC and baseline. The higher diversity of heatmaps in GAC shows the variability of parameters in GAC across groups.} 
	    \label{fig:8groupsheatmap}
	    \figvspace
\end{figure*}

\begin{table}[t]
    \centering
    \scalebox{0.62}{
    \begin{tabularx}{1.6\linewidth}{X c | X c c c c}
        \toprule
        \multirow{2}{*}{Method} & \multirow{2}{*}{LFW (\%)} & \multirow{2}{*}{Method} & IJB-A (\%) & \multicolumn{3}{c}{IJB-C @ FAR (\%)}\\
        \cline{5-7}
        & & & 0.1\% FAR & 0.001\% & 0.01\% & 0.1\% \\
        \midrule  
        DeepFace+~\cite{taigman2014deepface} & $97.35$ & Yin~\etal~\cite{yin2017multi} & $73.9 \pm 4.2$ & - & - & 69.3 \\
        CosFace~\cite{wang2018cosface} & $99.73$ & Cao~\etal~\cite{cao2018vggface2} & $90.4 \pm 1.4$ & $74.7$ & $84.0$ & $91.0$ \\
        ArcFace~\cite{Deng_2019_CVPR} & $\mathbf{99.83}$ & Multicolumn~\cite{xie2018multicolumn} & $\mathit{92.0 \pm 1.3}$ & $\underline{77.1}$ & $\underline{86.2}$ & $\underline{92.7}$ \\
        PFE~\cite{shi2019probabilistic} & $\mathit{99.82}$ & PFE~\cite{shi2019probabilistic} & $\mathbf{95.3 \pm 0.9}$ & $\mathbf{89.6}$ & $\mathbf{93.3}$ & $\mathbf{95.5}$ \\
        \midrule
        Baseline & $99.75$ & Baseline & $90.2 \pm 1.1$ & $80.2$ & $88.0$ & $92.9$ \\
        GAC & $\underline{99.78}$ & GAC & $\underline{91.3 \pm 1.2}$ & $\mathit{83.5}$ & $\mathit{89.2}$ & $\mathit{93.7}$\\
        \bottomrule
    \end{tabularx}}
        \caption{ \footnotesize{Verification performance on LFW, IJB-A, and IJB-C. [Key: \textbf{Best}, \textit{Second}, \underline{Third} Best]}}
        \label{tab:lfw_ijba_ijbc}
    \tablespace
\end{table}

\SubSection{Results on Standard Benchmark Datasets}
While our GAC mitigates bias, we also hope it can perform well on standard benchmarks.
Therefore, we evaluate GAC on standard benchmarks without considering demographic impacts, including LFW~\cite{huang2008labeled}, IJB-A~\cite{klare2015pushing}, and IJB-C~\cite{maze2018iarpa}. 
These datasets exhibit imbalanced distribution in demographics. For a fair comparison with SOTA, instead of using ground truth demographics, we train GAC on Ms-Celeb-1M~\cite{guo2016ms} with the demographic attributes estimated by the classifier pre-trained in Sec.~\ref{sec:gender_race}.
As in Tab.~\ref{tab:lfw_ijba_ijbc}, GAC outperforms the baseline and performs comparable to SOTA.

\SubSection{Visualization and Analysis on Bias of FR}
\textbf{Visualization:} To understand the adaptive kernels in GAC, we visualize the feature maps at an adaptive layer for faces of various demographics, via a Pytorch visualization tool~\cite{uozbulak_pytorch_vis_2019}.
We visualize important face regions pertaining to the FR decision by using a gradient-weighted class activation mapping (Grad-CAM)~\cite{selvaraju2017grad}. 
Grad-CAM uses the gradients back from the final layer corresponding to an input identity, and guides the target feature map to highlight import regions for identity predicting. 
Figure~\ref{fig:8groupsheatmap} shows that, compared to the baseline, the salient regions of GAC demonstrate more diversity on faces from different groups. This illustrates the variability of network parameters in GAC across different groups. 

\textbf{Bias via local geometry:} 
In addition to STD, we also explain the bias phenomenon by using the local geometry of a given face representation in each demographic group. 
We assume that the statistics of neighbors of a given point (representation) reflects certain properties of its manifold (local geometry). 
Thus, we illustrate the pair-wise correlation of face representations. 
To minimize variations caused by other latent variables, we use constrained frontal faces of a mug shot dataset, PCSO~\cite{klare2012face}, to show the demographic impact on face features. 
We randomly select $1$K White and $1$K Black subjects from PCSO, and compute their pair-wise correlation within each race. 
In Fig.~\ref{fig:pcso_histogram}, we discover that Base-White representations have lower inter-class correlation than Base-Black, {\it i.e.}, faces in the White group are over-represented by the baseline than faces in the Black group. 
In contrast, GAC-White and GAC-Black shows more similarity in their correlation histograms.

\begin{table}[t]
    \centering
    \tablespace
    \scalebox{0.7}{
    \begin{tabular}{cc ccc ccc cc}
        \toprule
        \multirow{2}{*}{Race} && \multicolumn{2}{c}{Mean} && \multicolumn{2}{c}{StaD} &&  \multicolumn{2}{c}{Relative Entropy}\\
        \cline{3-4} \cline{6-7} \cline{9-10}
        && Baseline & GAC && Baseline & GAC && Baseline & GAC \\
        \midrule
        White && $1.15$ & $1.17$ && $0.30$ & $0.31$ && $0.0$ & $0.0$ \\ 
        Black && $1.07$ & $1.10$ && $0.27$ & $0.28$ && $0.61$ & $0.43$ \\
        East Asian && $1.08$ & $1.10$ && $0.31$ & $0.32$ && $0.65$ & $0.58$ \\
        South Asian && $1.15$ & $1.18$ && $0.31$ & $0.32$ && $0.19$ & $0.13$ \\
        \bottomrule
    \end{tabular}}
        \caption{\footnotesize{Distribution of ratios between minimum inter-class distance and maximum intra-class distance of face features in $4$ race groups of RFW. GAC exhibits higher ratios, and more similar distributions to the reference.}}
    \label{tab:bias_distribute}
    \tablespace
\end{table}

As PCSO has few Asian subjects, we use RFW to design a second way to examine the local geometry in $4$ race groups.
That is, after normalizing the representations, we compute the pair-wise Euclidean distance and measure the ratio between the minimum distance of inter-subjects pairs and the maximum distance of intra-subject pairs.
We compute the mean and standard deviation (StaD) of ratio distributions in $4$ groups, by two models.
Also, we gauge the relative entropy to measure the deviation of distributions from each other. 
For simplicity, we choose White group as the reference distribution.
As shown in Tab.~\ref{tab:bias_distribute}, while GAC has minor improvement over baseline in the mean, it gives smaller relative entropy in the other $3$ groups,  indicating that the ratio distributions of other races in GAC are more similar, {\it i.e.}, less biased, to the reference distribution.
These results demonstrate the capability of GAC to increase fairness of face representations.

\Section{Conclusion}
This paper tackles the issue of demographic bias in face recognition by learning a fair face representation. A group adaptive classifier (GAC) is proposed to improve robustness of representations for every demographic group. Both adaptive convolution kernels and channel-wise attention maps are introduced to GAC. 
We further add an automation module to determine whether to use adaptations in a given layer. Our findings suggest that faces can be better represented by using layers adaptive to different demographic groups, leading to more balanced performance gain for all groups. 

{\small
\bibliographystyle{ieee_fullname}
\bibliography{egbib}
}

\end{document}


\title{Mitigating Face Recognition Bias via Group Adaptive Classifier \\ (Supplementary Material)}

\author{Sixue Gong\quad\quad Xiaoming Liu\quad\quad Anil K. Jain\\
Michigan State University, East Lansing MI 48824\\
{\tt\small \{gongsixu, liuxm, jain\}@msu.edu}
}

\maketitle
\thispagestyle{empty}
In this supplementary material we include; (1) Section \ref{sec:datasets}: the statistics of datasets used in the experiments; (2) Section \ref{sec:classifier}: Performance of the pre-trained gender and race/ethnicity classifiers to provide GAC with demographic information; (3) Section \ref{sec:demog}: the study on demographic proportions in training set and the intrinsic bias.

\section{Datasets}
\label{sec:datasets}

\begin{table*}[h]
    \centering
    \captionsetup{font=footnotesize}
    \caption{Statistics of training and testing datasets for the experiments in the paper}
    \label{tab:datasets}
    \scalebox{1.0}{
    \begin{tabularx}{0.9\linewidth}{c c c c c c c}
        \toprule
        Datasets && \# of Images && \# of Subjects && Demographic Annotations \\
        \midrule  
        IMDB~\cite{rothe2018deep} && $460,723$ && $20,284$ && Gender, Age \\
        UTKFace~\cite{zhifei2017cvpr} && $24,106$ && - && Gender, Age, Race/ethnicity \\
        AgeDB~\cite{moschoglou2017agedb} && $16,488$ && $567$ && Gender, Age \\
        AFAD~\cite{niu2016ordinal} && $165,515$ && - && Gender, Age, Ethnicity (East Asian) \\
        AAF~\cite{cheng2019exploiting} && $13,322$ && $13,322$ && Gender, Age \\
        RFW~\cite{wang2019racial} && $665,807$ && - && Race/Ethnicity \\
        BUPT-Balancedface~\cite{wang2020mitigating} && $1,251,430$ && $28,000$ && Race/Ethnicity\\
        IMFDB-CVIT~\cite{imfdb} && $34,512$ && $100$ && Gender, Age Groups, Ethnicity (South Asian) \\
        MS-Celeb-1M~\cite{guo2016ms} && $5,822,653$ && $85,742$ && No Demographic Labels \\
        PCSO~\cite{deb2017face} && $1,447,607$ && $5,749$ && Gender, Age, Race/Ethnicity \\
        LFW~\cite{huang2008labeled} && $13,233$ && $5,749$ && No Demographic Labels \\
        IJB-A~\cite{klare2015pushing} && $25,813$ && $500$ && Gender, Age, Skin Tone \\
        IJB-C~\cite{maze2018iarpa} && $31,334$ && $3,531$ && Gender, Age, Skin Tone \\
        \bottomrule
    \end{tabularx}}
\end{table*}

Tab.~\ref{tab:datasets} summarizes the datasets we adopt for conducting experiments, which reports the total number of face images and subjects (identities), and the types of demographic annotations. In the cross-validation experiment in Tab.~\ref{tab:cross_valid}, we report the statistics of each data fold for the cross-validation experiment on BUPT-Balancedface and RFW datasets.

\begin{table*}[h]
    \centering
    \captionsetup{font=footnotesize}
    \caption{Statistics of Dataset Folds in the Cross-validation Experiment}
    \label{tab:cross_valid}
    \scalebox{1.0}{
    \begin{tabular}{cc ccc ccc ccc cc}
        \toprule
        \multirow{2}{*}{Fold} && \multicolumn{2}{c}{White (\#)} && \multicolumn{2}{c}{Black (\#)} && \multicolumn{2}{c}{East Asian (\#)} && \multicolumn{2}{c}{South Asian (\#)} \\
        \cline{3-4} \cline{6-7} \cline{9-10} \cline{12-13}
        && Subjects & Images && Subjects & Images && Subjects & Images && Subjects & Images \\ 
        \midrule
        $1$ && $1,991$ & $68,159$ && $1,999$ & $67,880$ && $1,898$ & $67,104$ && $1,996$ & $57,628$ \\
        $2$ && $1,991$ & $67,499$ && $1,999$ & $65,736$ && $1,898$ & $66,258$ && $1,996$ & $57,159$ \\
        $3$ && $1,991$ & $66,091$ && $1,999$ & $65,670$ && $1,898$ & $67,696$ && $1,996$ & $56,247$ \\
        $4$ && $1,991$ & $66,333$ && $1,999$ & $67,757$ && $1,898$ & $65,341$ && $1,996$ & $57,665$ \\
        $5$ && $1,994$ & $68,597$ && $1,999$ & $67,747$ && $1,898$ & $68,763$ && $2,000$ & $56,703$ \\
        \bottomrule
    \end{tabular}}
\end{table*}

\section{Demographic Attribute Estimation}
\label{sec:classifier}

\begin{figure}[t]
    \captionsetup{font=footnotesize}
    \centering
    \begin{subfigure}[b]{0.53\linewidth}
    \includegraphics[width=\linewidth]{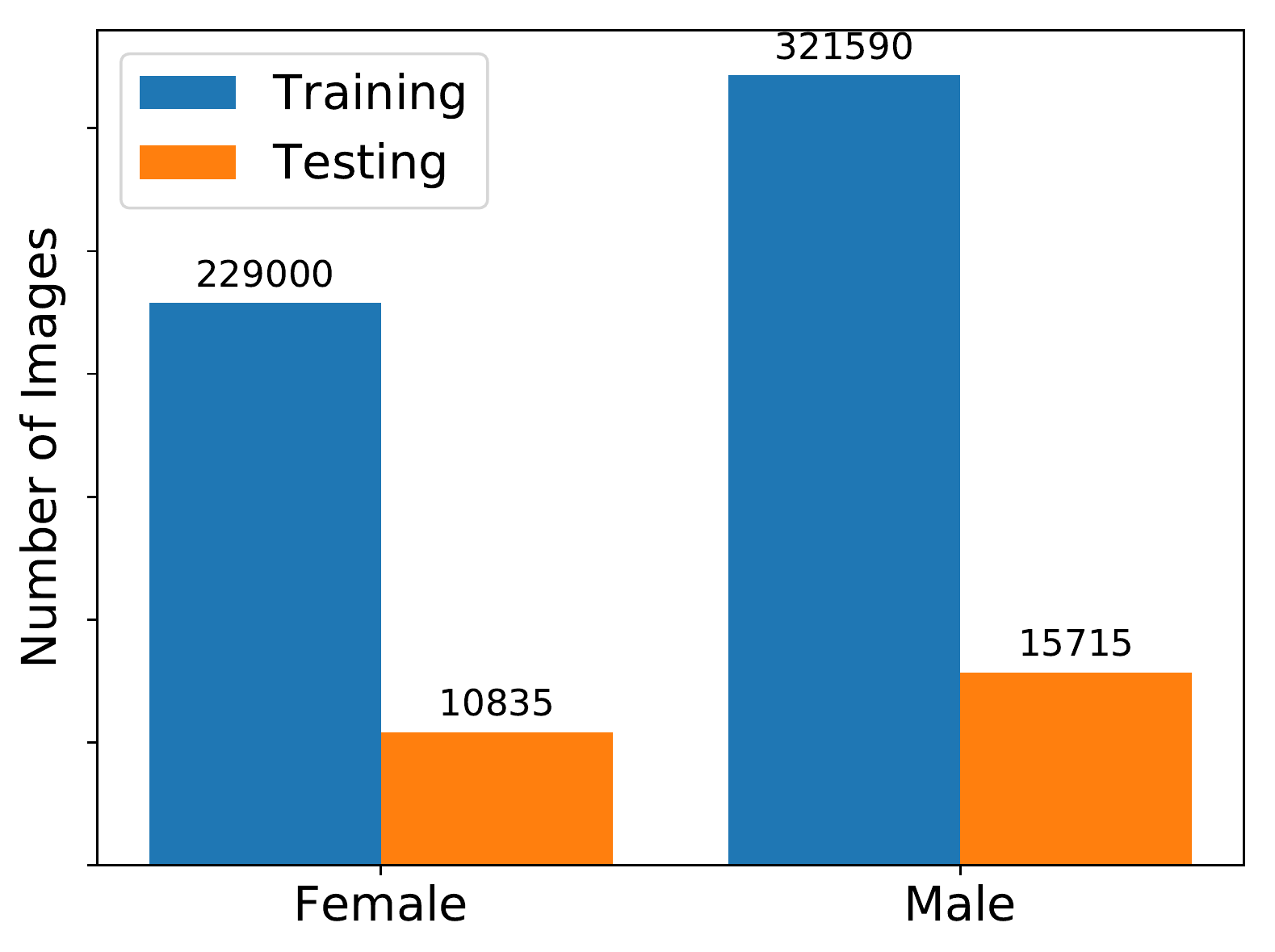}
    \vspace{-6mm}
    \caption{{\footnotesize} Gender Distribution}
    \end{subfigure}\hfill
    \begin{subfigure}[b]{0.46\linewidth}
    \includegraphics[width=\linewidth]{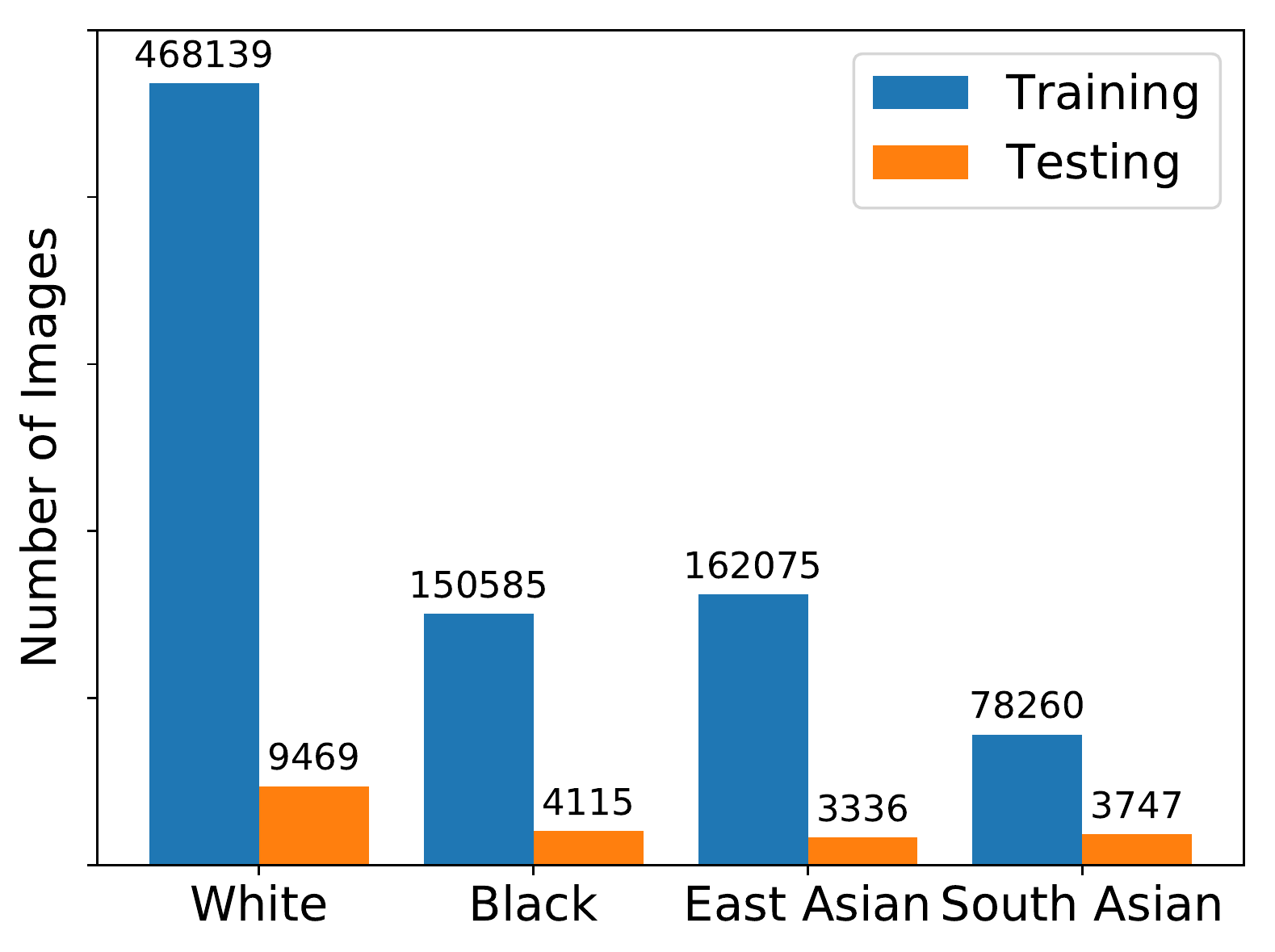}
     \vspace{-2mm}
    \caption{{\footnotesize} Race Distribution} 
    \end{subfigure}\\	  
    \vspace{-1mm}
    \caption{\footnotesize{Statistics of the datasets for training and testing demographic attribute estimation networks. (a) The number of images in each gender group of the datasets for gender estimation; (b) The number of images in each race/ethnicity group of the datasets for race/ethnicity estimation.} \vspace{-4mm}}
    \label{fig:stats}
\end{figure}

\begin{figure}[t]
    \captionsetup{font=footnotesize}
    \centering
    \begin{subfigure}[b]{0.53\linewidth}
    \includegraphics[width=\linewidth]{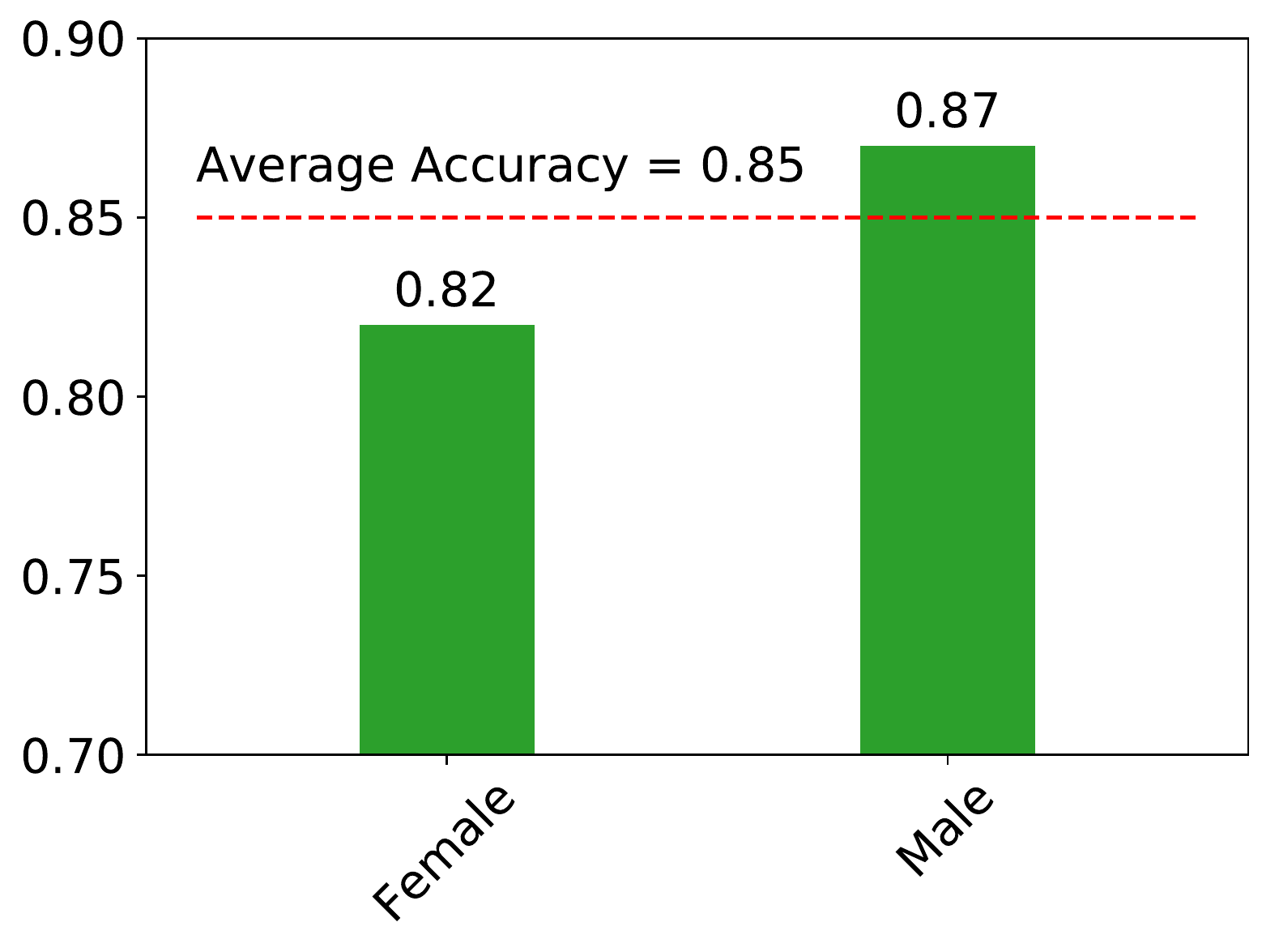}
    \vspace{-6mm}
    \caption{{\footnotesize} Gender Estimation}
    \end{subfigure}\hfill
    \begin{subfigure}[b]{0.46\linewidth}
    \includegraphics[width=\linewidth]{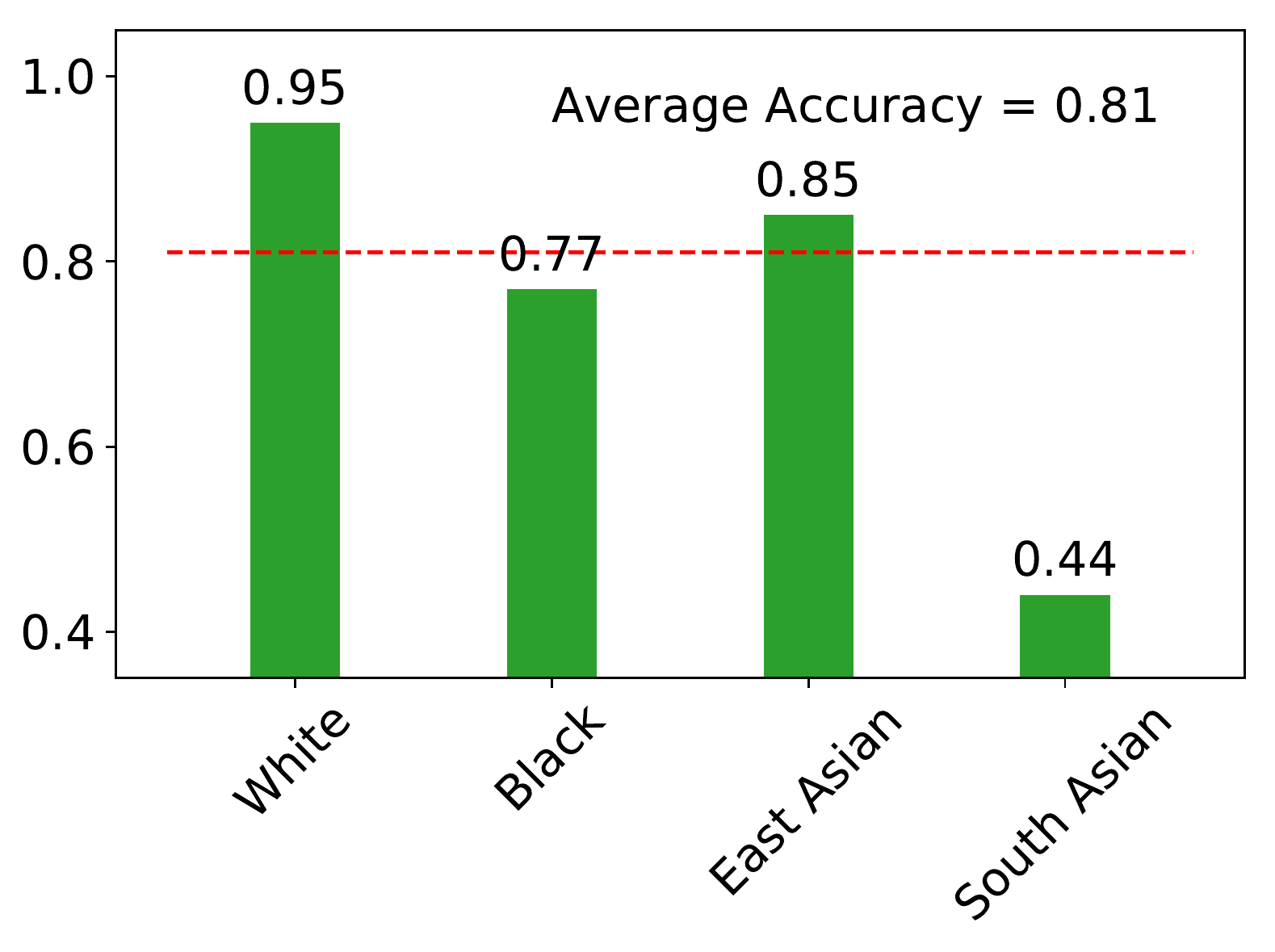}
     \vspace{-2mm}
    \caption{{\footnotesize} Race/Ethnicity Estimation} 
    \end{subfigure}\\	  
    \vspace{-1mm}
    \caption{\footnotesize{Performance of the demographic attribute estimation networks. (a) The classification accuracy in each gender group; (b) The classification accuracy in each race/ethnicity group. The red dashed line shows the average performance.} \vspace{-4mm}}
    \label{fig:acc}
\end{figure}

We train a gender classifier and a race/ethnicity classifier to provide GAC with demographic information during both training and testing procedures. We use the same datasets for training and evaluating the two demographic attribute classifiers as the work of~\cite{gong2020jointly}. The combination of IMDB, UTKface, AgeDB, AFAD, and AAF is used for gender estimation, and the collection of AFAD, RFW, IMFDB-CVIT, and PCSO is used for race/ethnicity estimation. Fig.~\ref{fig:stats} shows the total number of images in each demographic group of the training and testing set. Fig.~\ref{fig:acc} shows the performance of demographic attribute estimation on the testing set. For gender estimation, we see that the performance in the male group is better than that in the female group. For race/ethnicity estimation, the white group outperforms the other race/ethnicity groups.

\begin{table}[t]
    \captionsetup{font=footnotesize}
    \centering
    \caption{\footnotesize Verification accuracy ($\%$) on the RFW protocol~\cite{wang2019racial} with varying race/ethnicity distribution.}
    \label{tab:bias_rfw_ratio}
   \scalebox{0.7}{
    \begin{tabular}{@{}c cccc cc@{}}
        \toprule
        Training Ratio & White & Black & East Asian & South Asian & Avg ($\uparrow$) & STD ($\downarrow$)\\
        \midrule
        7:7:7:7 & $96.20$ & $94.77$ & $94.87$ & $94.98$ & $95.21$ & $0.58$\\
        5:7:7:7 & $96.53$ & $94.67$ & $94.55$ & $95.40$ & $95.29$ & $0.79$ \\
        3.5:7:7:7 & $96.48$ & $94.52$ & $94.45$ & $95.32$ & $95.19$ & $0.82$ \\
        1:7:7:7 & $95.45$ & $94.28$ & $94.47$ & $95.13$ & $94.83$ & $0.48$ \\
        0:7:7:7 & $92.63$ & $92.27$ & $92.32$ & $93.37$ & $92.65$ & $0.44$ \\
        \bottomrule
    \end{tabular}}
\end{table}

\section{Analysis on Intrinsic Bias and Data Bias}
\label{sec:demog}
For all the algorithms listed in Tab.~1 of the main paper, the performance is higher in White group than those in the other three groups, even though all the models are trained on a demographic balanced dataset, BUPT-Balancedface~\cite{wang2020mitigating}. In this section, we further investigate the intrinsic bias of face recognition between demographic groups and the impact of the data bias in the training set. \textit{Are non-White faces inherently difficult to be recognized for existing algorithms? Or, are face images in BUPT-Balancedface (the training set) and RFW~\cite{wang2019racial} (testing set) biased towards the White group?}

To this end, we train our GAC network using training sets with different race/ethnicity distributions and evaluate them on RFW. In total, we conduct four experiments, in which we gradually reduce the total number of subjects in the White group from the BUPT-Balancedface dataset. To construct a new training set, subjects from the non-White groups in BUPT-Balancedface remain the same, while a subset of subjects is randomly picked from the White group. As a result, the ratios between non-White groups are consistently the same, and the ratios of White, Black, East Asian, South Asian are $\{5:7:7:7\}$, $\{3.5:7:7:7\}$, $\{1:7:7:7\}$, $\{0:7:7:7\}$ in the four experiments, respectively. In the last setting, we completely remove White from the training set.

Tab.~\ref{tab:bias_rfw_ratio} reports the face verification accuracy of models trained with different race/ethnicity distributions on RFW. For comparison, we also put our results on the balanced dataset here (with ratio $\{7:7:7:7\}$), where all images in BUPT-Balancedface are used for training. From the results, we see several observations: (1) It shows that the White group still outperforms the non-White groups for all the first three experiments. Even without any White subjects in the training set, the accuracy on the White testing set is still higher than those on the testing images in Black and East Asian groups. This suggests that White faces are either intrinsically easier to be verified or face images in the White group of RFW are less challenging. (2). With the decline in the total number of White subjects, the average performance declines as well. In fact, for all these groups, the performance suffers from the decrease in the number of White faces. This indicates that face images in the White groups are helpful to boost the performance of face recognition algorithms for faces from both White and non-White groups. In other words, faces from the White group benefit the representation learning of global patterns for face recognition in general. (3). Opposite to our intuition, the biasness is lower with less number of White faces, while the data bias is actually increased by adding the unbalancedness to the training set.

{\small
\bibliographystyle{ieee_fullname}
\bibliography{egbib}
}